\documentclass[11pt]{article}
\usepackage[letterpaper,margin=1in]{geometry}
\usepackage[round]{natbib}

\usepackage{amsmath,amssymb,amsthm,mathtools}
\usepackage{booktabs}
\usepackage[ruled]{algorithm}
\usepackage{algpseudocode}
\usepackage[section]{placeins}
\usepackage[hidelinks]{hyperref}
\usepackage{url}

\newtheorem{assumption}{Assumption}
\newtheorem{proposition}{Proposition}
\newtheorem{theorem}{Theorem}
\newtheorem{corollary}{Corollary}
\newtheorem{lemma}{Lemma}
\newtheorem{example}{Example}
\newtheorem{definition}{Definition}
\theoremstyle{remark}
\newtheorem{remark}{Remark}

\DeclareMathOperator*{\argmax}{arg\,max}
\newcommand{\E}{\mathbb{E}}
\newcommand{\Prob}{\mathbb{P}}
\newcommand{\cC}{\mathcal{C}}
\newcommand{\cI}{\mathcal{I}}
\newcommand{\cK}{\mathcal{K}}
\newcommand{\cX}{\mathcal{X}}
\newcommand{\Yc}{\tilde Y}

\usepackage{graphicx}
\title{The Staged Pandora's Box: Reservation Indices for Multi-Fidelity Bayesian Optimization}
\author{Long Vu\textsuperscript{1,2} \quad Jongseong Brad Choi\textsuperscript{1,3}\\[0.25em]
Binoy Ravindran\textsuperscript{2} \quad Zhoulai Fu\textsuperscript{1,2,3}\\[0.5em]
{\small \textsuperscript{1}State University of New York Korea}\\
{\small \textsuperscript{2}Virginia Tech \quad \textsuperscript{3}Stony Brook University}\\
{\small \texttt{long-vu@outlook.com}}}
\hypersetup{pdfauthor={Long Vu, Jongseong Brad Choi, Binoy Ravindran, Zhoulai Fu}}
\date{}
\begin{document}
\maketitle
\begin{abstract}
A staged search must decide which candidate to advance next and when to stop. Some observations
only inform later decisions; others can be kept as results. We formulate these choices as a
staged Pandora's box with a designated set of collectable stages. For independent finite-stage
candidates with bounded readings and positive bounded own-history costs, an exact covered-call recursion gives the
reservation index, the optimal value and the loss of every advance or stop; advancing an
index-maximal candidate while its index exceeds the incumbent is optimal. The representation
recovers classical single-stage search, nested inspection and the all-collectable Markov setting,
and prices changes in the problem: enlarging the collectable set or lowering costs raises the
value by an explicit covered-call overshoot. Resumption can be valuable even when future
reward laws do not change, but has no value for deterministic readings. With information
sharing across candidates, no rule using only their
separate remaining-reading laws is universally optimal, and making an intermediate reading
collectable can change which candidate is inspected first. Under a coherent joint path law
with own-history costs, an approximate-planning bound controls the recomputed index's regret
by changes in untouched candidates' local values; a common-component
specialization bounds those changes through predictive total-variation dispersion.
Exact finite-law experiments check the representation, the loss
accounting and the bounds.
\end{abstract}

\section{Introduction}
\label{sec:intro}

Cost-aware search has two coupled decisions, in Weitzman's terms a \emph{selection rule},
which candidate to inspect next, and a \emph{stopping rule}, whether to inspect at all
\citep{weitzman1979}. For independent candidates inspected in one step, one reservation value
answers both: open a box of largest reservation value while that value exceeds the best
result already available. The Pandora's box Gittins index (PBGI) brings this principle to
cost-aware Bayesian optimization \citep{xie2024pbgi,xie2025coststop}.

An evaluation need not reveal its result at once. A training run may be paused at checkpoints
and resumed later; a proxy evaluation may inform the decision to pay for a full one without
being a deployable result; a learning curve may make every checkpoint a usable result. We
represent these situations with a finite ladder per candidate and a \emph{collectable set}:
only readings at designated stages enter the incumbent. The searcher may advance any available
candidate by one stage, return to another, or stop and keep its best collectable reading less
the costs incurred. Staging changes both rules: selection includes resuming a paused candidate,
and what stopping keeps depends on the collectable set.

In multi-fidelity Bayesian optimization, a candidate can be a fixed hyperparameter
configuration and its ladder a prescribed sequence of fidelities or training checkpoints.
Terminal-only collection models proxy evaluations whose outputs cannot be deployed;
checkpoint collection models intermediate outputs that are themselves admissible results.
We study the search structure on a finite candidate pool under bounded planning laws;
the numerical studies verify this structure on known laws.

We give an explicit covered-call representation of this problem. At a nonexhausted arm
state $s$ with a collectable stage left, let $\Yc$ be the next stage's reading if it
is collectable and $-\infty$ otherwise, and let $s'$ be the successor own state.
The recursion
\[
W(s)=\max\{\Yc,Z(s')\},\qquad Z(s)=\min\{G(s),W(s)\},\qquad
\E[(W(s)-G(s))^+]=c(s),
\]
represents each remaining arm by its next stage's collectable reading or future capped
value, capped again at its current reservation value $G(s)$. A suffix without a collectable
stage, including an exhausted arm, has $G=Z=-\infty$ and needs no reservation equation.
Under independent arms the optimal value is the expectation of the largest
covered call value and the incumbent (Theorem~\ref{thm:index}). The same representation gives
the exact Bellman loss of every advance and stop, hence of any policy
(Theorem~\ref{thm:loss}), and prices changes in the problem: enlarging the collectable set or
lowering costs raises the value by an explicit overshoot, planning with a smaller collectable
set loses at most that amount, and deterministic readings need no resumption.
With stochastic collectable rewards, resumption can help without learning about
unobserved rewards (Section~\ref{sec:comparative}).

We do not claim a new index theorem: the policy is consistent with classical superprocess
results \citep{whittle1980gittins,glazebrook1982whittle}, and the single-stage, terminal-only
and all-collectable Markov cases have established solutions (Section~\ref{sec:related}). Our
contribution is the explicit recursion, value representation and action losses for designated
collection, together with the effects of collection rights on acquisition and the local-value
discrepancy under sharing. The construction separates information from collectable reward
while accommodating own-history states and costs, features already permitted in nested
inspection models \citep[Section~4.1]{bowers2024nested}.

Under a shared surrogate, one candidate's observation changes the law of the others, and
separate arm problems omit the value of that change. Two instances with identical separate
arm laws can have different uniquely optimal first actions (Proposition~\ref{prop:no-local}),
and collection rights reshape acquisition under sharing: in a two-arm family with a common
latent state, the optimal first advance alternates $A,B,A,B$ in the prior when the second
arm's intermediate reading is collectable, and is always $A$ when it is not
(Proposition~\ref{prop:switching}). For a coherent joint path law with arm-local costs, an
approximate-planning bound controls the recomputed index's regret by how much untouched arms'
local values change (Proposition~\ref{prop:sharing-robustness}); when sharing flows through a
common component, a local envelope is the arms' sensitivity to the component times the
predictive total-variation dispersion of one more reading (Corollary~\ref{cor:component}).
A vanishing regret bound requires this envelope to vanish uniformly over feasible future
histories; posterior concentration alone is insufficient. These conditions are not
established for a general correlated Gaussian process.

\subsection{Contributions}
\label{sec:contributions}
We provide (i) the staged model and a covered-call recursion giving the optimal value and
index policy under independence; (ii) exact stop and advance losses with an arbitrary-policy
decomposition, exact comparative statics for collection rights and costs, and a bound on the
value of resumption; (iii) under information sharing, the limits of own-arm-law policies, a
certified family in which collection rights split the optimal first-action region, and an
approximate-planning bound expressed through local-value changes and a common-component
dispersion envelope; and (iv) exact finite-state checks of the value and loss identities,
comparative statics and sharing bounds against direct Bellman optimization.

\subsection{Related work}
\label{sec:related}

Multi-stage inspection has a reservation-index theory. \citet[Appendix~G]{kleinberg2016descending}
give generalized strike prices and \citet[Section~4, Definition~4]{bowers2024nested} a
constructive nested-index and capped-value recursion, both for terminal-only collection;
the latter also explicitly allow later costs to depend on preceding signals
\citep[Section~4.1]{bowers2024nested}.
\citet{gupta2019markovian} treat independent Markov inspection processes with rewards at
destination states, nonnegative prices and unbounded depth, of which the fixed-depth
positive-cost case is recovered here; \citet{boodaghians2023order} handle independent rewards
under line or tree precedence. Closest for intermediate collection is
\citet[Theorem~4.4]{yang2025markovian}: each reward on a Markov line is collectable, probing
costs are fixed, and the optimal policy selects the line with largest generalized reservation
value. With positive costs that is the corresponding special case here; their finite-support
dynamic program and our history-state covered-call construction are different
representations. The superprocess literature supplies sufficient conditions for an index
theorem \citep{whittle1980gittins,glazebrook1982whittle,scully2025gittins};
Appendix~\ref{app:remarks} identifies that condition for this model. More general Markov
search, in which a process offers a choice of actions, need not admit an index policy
\citep{glazebrook1982whittle,scully2025gittins}; one-shot search with an optional partial
inspection has none and is NP-hard \citep{aouad2025sequential}, as is combinatorial Markov
search \citep{bowers2026cms}.

Correlated search is also established: learning and approximation from joint-distribution
samples \citep{chawla2020correlations}, the reduction of fully adaptive correlated search to
Uniform Decision Tree \citep{chawla2023adaptivecorrelations}, correlation-aware reservation
rules \citep{gergatsouli2023weitzman}, and a tight factor four against the best partially
adaptive policy for a minimization objective \citep{bansal2025correlated}. Those approximation
factors do not transfer to our reward-minus-cost objective, and the correlated-BO stopping
guarantee of \citet{xie2025coststop}, no worse than stopping after one evaluation, is a
different kind of statement from Bellman optimality. In Bayesian optimization, PBGI is the
single-stage predecessor \citep{xie2024pbgi}; freeze-thaw and fidelity-aware surrogates
\citep{swersky2014,klein2017fabolas} motivate staged observation models; trace-aware knowledge
gradient \citep{wu2019takg}, successive halving and Hyperband
\citep{jamieson2016sh,li2018hyperband} allocate resources by other means. For this utility in
fixed-pool freeze-thaw optimization, \citet{vu2026coststop} analyze the one-step rule, expected
improvement stopping, and a certified two-step correction; the present paper gives the exact
index for the independent-arm staged problem.

\paragraph{Organization.}
Section~\ref{sec:setting} defines the model. Section~\ref{sec:main} states the
covered-call recursion, the index theorem and the loss decomposition, with
proofs; Sections~\ref{sec:comparative} and~\ref{sec:robust} treat comparative
statics and information sharing. Section~\ref{sec:validation} summarizes the
exact numerical verification, whose retained models, checks and results are in
Appendix~\ref{app:experiments}, and Section~\ref{sec:limitations} concludes.
Algorithm~\ref{alg:staged-index} gives the policy implementation in the main text.

\section{The Staged Pandora's Box}
\label{sec:setting}

\paragraph{Candidates, ladders and the collectable set.}
There is a fixed pool $\cX=\{x_1,\ldots,x_N\}$ of candidates, the \emph{arms}. Arm $i$ has a fixed finite
ladder of $T_i$ stages, or rungs; after $b$ advances in total it stands at stage $t_{i,b}\in\{0,\ldots,T_i\}$,
and one action, an \emph{advance}, takes it to stage $t_{i,b}+1$ at a cost and reveals a
reading $Y$. The set $\cK_i\subseteq\{1,\ldots,T_i\}$ is fixed before the run. A stage
$t\in\cK_i$ is \emph{collectable}: its reading
enters the incumbent. Write $\Yc=Y$ if the stage reached is collectable and $\Yc=-\infty$
otherwise, the \emph{collectable reading}. The history state is
$S_b=(\cC_b,t_{1:N,b},\tilde y_b,K_b)$. Here $\cC_b$ records the ordered action and observation
history, including noncollectable readings and any fixed starting context; $t_{1:N,b}$ records
the arms' stages; $\tilde y_b$ is the best collectable reading seen or the initial value $q_0$;
and $K_b$ is cumulative cost. The \emph{planning law}
$p_\theta(\cdot\mid S_b,i)$ is the predictive distribution of arm $i$'s next reading under a
specified predictive model whose parameters $\theta$ are fixed before the run and never refitted; it may condition
on the whole context, so an observation on one arm can change the prediction for every arm.
Advancing $i\in\cI(S_b)$, the set of arms with a stage left, costs $c(S_b,i)$ and reveals
$Y_{b+1}\sim p_\theta(\cdot\mid S_b,i)$; the successor $S_{b+1}=F(S_b,i,Y_{b+1})$ adds the
observation to the context, advances $t_i$ by one, sets
$\tilde y_{b+1}=\max\{\tilde y_b,\Yc_{b+1}\}$ and adds the cost to $K_b$. Availability is
arm-local: advancing $i$ never changes another arm's ladder. A policy $\pi$ chooses, measurably
in the observed history, an action in $\{\mathrm{stop}\}\cup\cI(S_b)$; $\tau_\pi$ is its
stopping time. Randomized policies may use auxiliary randomness independent of the innovations
driving the planning kernel. The searcher always retains earlier collectable readings, so collecting at an
intermediate stage does not require terminating the search.

\paragraph{Utility, values and the Bellman equation.}
The objective is $\sup_\pi\E_{\theta,\pi}[\tilde y_{\tau_\pi}-K_{\tau_\pi}\mid S_0]$. It is the
budget-free utility of \citet{vu2026coststop} for fixed-pool freeze-thaw optimization, with a
cost per stage in place of a cost per increment and only collectable readings entering the
score. The stop-now payoff is $h(S_b)=\tilde y_b-K_b$, the value of $\pi$ is
$V_\pi(S_b)=\E_{\theta,\pi}[\tilde y_{\tau_\pi}-K_{\tau_\pi}\mid S_b]$, and
$Q_\pi(S_b,i)=\E_\theta[V_\pi(F(S_b,i,Y_{b+1}))\mid S_b,i]$ is the value of forcing $i$ once and
then following $\pi$. The optimal value $V^\star=\sup_\pi V_\pi$ satisfies
$V^\star(S)=\max\{h(S),\max_{i\in\cI(S)}Q^\star(S,i)\}$, with $\max\varnothing=-\infty$, and the
ladders being finite, a policy that attains the maximum at every state is optimal.
For comparison with continuation-value conventions, define
$U_\pi(S)=V_\pi(S)+K(S)$. With $S'=F(S,i,Y)$,
\[
 Q_\pi(S,i)+K(S)=-c(S,i)+\E_\theta[U_\pi(S')\mid S,i].
\]
Thus $V_\pi$ includes sunk costs, while $U_\pi$ removes them; the cost of the
forced advance is already included in $K(S')$ inside the definition of $Q_\pi$.

\begin{assumption}[Standing assumptions]\label{assm:standing}
The planning law is a fixed measurable kernel, never refitted during a run; all readings and
incumbents lie in fixed bounds $[y_{\min},y_{\max}]$ chosen in advance; the cost $c(S,i)$ is measurable, determined by
$(S,i)$ before the advance, and satisfies $0<c_{\min}\le c(S,i)\le c_{\max}<\infty$; every ladder
is finite. History states carry the Borel sigma-fields generated by the finite arm labels and
their real observations; the successor map is measurable.
\end{assumption}

\paragraph{Finite horizon and conventions.}
We write $q(S)=\tilde y(S)$ and $L(S)=\sum_i(T_i-t_i)$ throughout. Every advance lowers $L$
by one, so every admissible policy stops after at most $L(S)$ further advances, and
finite-action backward induction defines Borel Bellman values and an attaining deterministic
policy, choosing the first action in any fixed measurable tie order. The stated index policy
stops on an index--incumbent tie. We use $(-\infty-r)^+=0$ for finite $r$ and
$\max\varnothing=\sup\varnothing=-\infty$. Appendix~\ref{app:notation} collects the remaining
conventions and compares them with the closest index and inspection models.

\paragraph{Per-arm planning laws.}
A planning law is \emph{per-arm} when there are measurable kernels $P_i(dy\mid s_i)$ and
cost functions $c_i(s_i)$ such that $p_\theta(dy\mid S,i)=P_i(dy\mid s_i)$ and
$c(S,i)=c_i(s_i)$. An arm's \emph{own state} $s_i$ is its stage and complete observation
history. Advancing another arm leaves $s_i$, its predictive kernel and its next cost unchanged;
an arm does not evolve while paused. We write $\nu(s_i)=T_i-t_i$ and abbreviate its cost as
$c(s_i)$. No Gaussian, density, or scalar first-order Markov assumption is imposed.

At a fixed state $S$, iterate $P_i$ to obtain the law $\mathsf P_i^{s_i}$ of the entire
remaining path of arm $i$. The common space used in the independent-arm proofs has law
\[
\mathsf P_S=\bigotimes_{i=1}^N\mathsf P_i^{s_i}.
\]
One can sample these paths in advance and reveal their next coordinates as the policy chooses
arms. Autonomy makes this construction agree with the sequential planning kernels. Revealing a
coordinate of one path conditions that path alone; the other remaining paths keep their laws.
Dependence within a path is unrestricted and its future costs are functions of its successive
own states. Theorem~\ref{thm:index} uses this product construction; Section~\ref{sec:robust}
does not assume it.

\paragraph{Single-arm stopping problems.}
Fix a history state $S$, an available arm $i$, and a depth cap $k\ge1$. Advance only $i$,
obtaining states $S^{(m)}$ by iterating the planning kernel and successor map. Let
$\mathcal G_m$ be the filtration of those observations and $\nu_i=T_i-t_i$ the remaining
stages. An admissible arm run is a stopping time $1\le\sigma\le\min\{k,\nu_i\}$ for this
filtration; it pays every advance up to $\sigma$ and may abandon the arm after any reading.
Write $\mathcal R_i^{(k)}(S)$ for these runs, $M_\sigma$ for the best newly obtained
collectable reading ($-\infty$ if none), and $\kappa_\sigma$ for the run's total
additional cost. Earlier collectable readings remain in the incumbent, rather
than in $M_\sigma$. For an outside value $r$, define
\[
\Phi_i^{(k)}(S;r)=\sup_{\sigma\in\mathcal R_i^{(k)}(S)}
\{\E[(M_\sigma-r)^+\mid S]-\E[\kappa_\sigma\mid S]\},\qquad
G_i^{(k)}(S)=\sup\{r:\Phi_i^{(k)}(S;r)>0\}.
\]
The empty supremum for $G$ is $-\infty$. The run law is the original planning law from $S$;
$r$ is a payoff comparison, not a modification of that law. The full index uses $k=\nu_i$;
exhausted arms have index $-\infty$. If a collectable stage remains, let $\mu_i(S)$ be the
number of advances to the next such stage. The first advance is forced in $\Phi$; the option
of stopping immediately is represented by $\Phi^+=\max\{0,\Phi\}$. The auxiliary value $r$
may lie outside the range of observed scores.

\begin{proposition}[Reservation-value properties]\label{prop:index}
Under Assumption~\ref{assm:standing}, $r\mapsto\Phi_i^{(k)}(S;r)$ is nonincreasing and
$1$-Lipschitz. The index is finite exactly when a collectable stage lies within the cap, in
which case
\[
y_{\min}-\mu_i(S)c_{\max}\le G_i^{(k)}(S)\le y_{\max}-c(S,i).
\]
In every case $\Phi_i^{(k)}(S;r)>0$ if and only if $r<G_i^{(k)}(S)$, and $G_i^{(k)}(S)$ is
nondecreasing in $k$. When the index is finite, $\Phi_i^{(k)}(S;G_i^{(k)}(S))=0$.
\end{proposition}
Appendix~\ref{app:proof-details} gives the proof.

The index is measurable: it is also the supremum of the rational $a$ for which
$\Phi_i^{(k)}(S;a)>0$. Thus selecting a smallest-numbered index maximizer defines a measurable
policy even for continuous or atomic reading laws.

These definitions also apply under a shared planning law: the current history fixes the
starting conditional law, and the hypothetical run observes only the chosen arm. The
independent-arm theorem requires the additional autonomy assumption given above.

\section{Main Results}
\label{sec:main}

\subsection{Covered-call recursion and optimal policy}
\label{sec:nested}

Throughout this section the planning law is per-arm and
Assumption~\ref{assm:standing} holds. Fix a nonexhausted arm and write $s$ for its own state, $Y$ for its
next reading, $\Yc$ for the collectable reading and $s'$ for its state after it. Write $\Phi(s;r)=\Phi_i^{(\nu_i)}(S;r)$ and $G(s)=G_i^{(\nu_i)}(S)$ for the single-arm quantities of Section~\ref{sec:setting}.
All expectations in the local calculation are under $\mathsf P_i^s$; after observing $Y$,
future expectations use the conditional continuation law from $s'$. The continuation
improvement is zero at an exhausted arm; the outside option is retained.
Lemma~\ref{lem:run-dp} justifies the following Bellman
recursion and its attaining stopping rules for arbitrary within-arm kernels.
After the forced first advance the run
either stops or continues from $s'$ against the incumbent $\max\{r,\Yc\}$, and
$(\max\{\Yc,M'\}-r)^+-(\Yc-r)^+=(M'-\max\{r,\Yc\})^+$ for the best later collectable reading
$M'$, so
\begin{equation}
\Phi(s;r)=\E\bigl[(\Yc-r)^+-c(s)+\Phi(s';\max\{r,\Yc\})^+\bigr],
\label{eq:phi-bellman}
\end{equation}
with $\Phi(s';\cdot)^+\equiv0$ if $\nu(s')=0$.

\begin{definition}[Nested value and covered call value]\label{def:nested}
For an arm state $s$ define, by backward recursion on $\nu(s)$, $Z(s)=-\infty$ if $\nu(s)=0$ and
otherwise
\[
W(s)=\max\{\Yc,\,Z(s')\},\qquad Z(s)=\min\{G(s),\,W(s)\}.
\]
$W(s)$ is a function of the arm's readings from $s$ onward; $Z(s)$ is its \emph{covered call value}.
These are random variables on the full remaining arm path, not readings disclosed by the
first advance. The index $G(s)$ is a number determined by the current state.
Finite backward recursion makes $W,Z$ jointly measurable in the starting state and continuation
path, using measurability of $G$ from Section~\ref{sec:setting}.
\end{definition}

When collection remains, the backward computation is: obtain the child caps,
form $W$, solve its reservation equation for $G$, and cap $W$ to obtain $Z$.
The lemma below shows that this computed root equals the stopping threshold
already defined in Section~\ref{sec:setting}, and that the cap accounts for
the advance cost in the keep-or-run value.

\begin{lemma}[Nested box]\label{lem:nested}
For every arm state $s$ with $\nu(s)\ge1$ and every $r\in\mathbb R$,
\[
\Phi(s;r)=\E\bigl[(W(s)-r)^+\bigr]-c(s),\qquad
\E\bigl[(W(s)-G(s))^+\bigr]=c(s)\ \text{ if }G(s)>-\infty,
\]
and $\E[\max\{r,Z(s)\}]=r+\Phi(s;r)^+$. If a collectable stage remains, the reservation
equation has a unique finite solution, and $W(s),Z(s)$ are bounded real random variables.
If none remains, $G(s)=W(s)=Z(s)=-\infty$ and $\Phi(s;r)=-c(s)$.
The local value therefore has Weitzman's positive-part form, with $G(s)$ the reservation
price and $Z(s)$ the covered call value \citep{kleinberg2016descending}.
\end{lemma}
\begin{proof}
Induction on $\nu(s)$. If $\nu(s')=0$ then $W(s)=\Yc$ and \eqref{eq:phi-bellman} is the first
claim. Otherwise the hypothesis at $s'$ gives $\Phi(s';r')^+=\E[\max\{r',Z(s')\}\mid Y]-r'$ with
$r'=\max\{r,\Yc\}$, and \eqref{eq:phi-bellman} becomes
\begin{align*}
\Phi(s;r)&=\E\bigl[(\Yc-r)^+ + \max\{r,\Yc,Z(s')\}-\max\{r,\Yc\}\bigr]-c(s)\\
&=\E\bigl[\max\{r,\Yc,Z(s')\}-r\bigr]-c(s)=\E\bigl[(W(s)-r)^+\bigr]-c(s),
\end{align*}
using $(\Yc-r)^+=\max\{r,\Yc\}-r$, which holds also for $\Yc=-\infty$.
If no collectable stage remains, all continuation covered calls are $-\infty$ and stopping
after the forced advance gives $\Phi=-c(s)$. Otherwise backward induction, together with
Proposition~\ref{prop:index}, bounds the finite $W,Z$ between
$y_{\min}-\nu(s)c_{\max}$ and $y_{\max}$. This makes their positive-part expectations finite.
Continuity and Proposition~\ref{prop:index} give
$\E[(W-G)^+]=c(s)$. The root is unique: for $a<b$ with $\E[(W-b)^+]>0$,
\[
\E[(W-a)^+]-\E[(W-b)^+]\ge(b-a)\Prob(W>b)>0.
\]
Two roots cannot therefore have the same strictly positive level $c(s)$; no density or
absence-of-atoms assumption is needed. For the covered-call identity, if $r<G(s)$ then
$(\min\{G,W\}-r)^+=(W-r)^+-(W-G)^+$, so $\E[\max\{r,Z\}]=r+\Phi(s;r)$ with $\Phi(s;r)>0$; if
$r\ge G(s)$ then $Z\le G\le r$ gives $\E[\max\{r,Z\}]=r$ and $\Phi(s;r)\le0$.
\end{proof}

The representation holds for every outside option while preserving the staged observation protocol.
In particular,
$Y$ and $Z(s')$ may be dependent; the proof integrates their actual joint law. The role of
collectable intermediate readings is exactly the maximum in $W$. No discretization is used
in this lemma; Example~\ref{ex:continuous-collection} gives a continuous dependent instance
with both stages collectable.

\begin{theorem}[Index optimality under a per-arm planning law]\label{thm:index}
Under Assumption~\ref{assm:standing} with a per-arm planning law, for every state $S$
with incumbent $q$, cumulative cost $K$ and arm states $s_1,\ldots,s_N$,
\[
V^\star(S)+K=\E\Bigl[\max\bigl\{q,\ \max_i Z_i(s_i)\bigr\}\Bigr],
\]
the expectation being over the independent covered call values of the arms. Stopping is optimal if and
only if $\max_i G_i(s_i)\le q$, and when $\max_i G_i(s_i)>q$ advancing any arm with the largest
index is optimal. In particular the index policy $\pi_{\rm IX}$, which stops when
$\max_i G_i(s_i)\le q$ and otherwise advances an index-maximal arm, has $V_{\pi_{\rm IX}}=V^\star$.
\end{theorem}
\begin{proof}
All expectations below are under the product remaining-path law $\mathsf P_S$ from
Section~\ref{sec:setting}. Let $\mathcal V(S)=\E[\max\{q,\max_i Z_i(s_i)\}]$.
We verify this candidate by showing that every advance has nonpositive Bellman
residual, with equality for a maximal-index advance above $q$; otherwise stopping
attains the value. Finite-horizon induction then identifies $\mathcal V$ with $V^\star+K$.
Then $\mathcal V\ge q$, with equality when $\max_iG_i(s_i)\le q$ because $Z_i\le G_i$.
If an available arm has $G_i=-\infty$, it has no collectable stage left. Its advance changes
neither the incumbent nor the other arms' states or laws, so its Bellman residual is
$\E[\mathcal V(S')]-c(s_i)-\mathcal V(S)=-c(s_i)<0$.

Now fix an available arm with finite index, abbreviate $W_i=W_i(s_i)$ and $G_i=G_i(s_i)$, and
let $R=\max\{q,\max_{j\ne i}Z_j(s_j)\}$, the outside option supplied by the incumbent
and the rest of the pool. It is random but independent of arm $i$'s entire remaining path.
After its first reading $Y$, the conditional law of that arm is the law from
$s_i'$, while every other path retains its law. Consequently, almost surely,
\[
\mathcal V(S')=\E\!\left[\max\{q,\Yc,Z_i(s_i'),\max_{j\ne i}Z_j(s_j)\}\mid Y\right].
\]
Taking the outer expectation gives $\E[\mathcal V(S')]=\E[\max\{R,W_i\}]$ by the tower
property; it does not assume $W_i$ is observed. By definition,
$\mathcal V(S)=\E[\max\{R,\min\{G_i,W_i\}\}]$. The pathwise identity
\[
\max\{R,W\}-\max\{R,\min\{G,W\}\}=(W-\max\{R,G\})^+
\]
therefore gives
\begin{align*}
\E[\mathcal V(S')]-c(s_i)-\mathcal V(S)
&=\E\bigl[(W_i-\max\{R,G_i\})^+\bigr]-c(s_i)\\
&\le\E\bigl[(W_i-G_i)^+\bigr]-c(s_i)=0
\end{align*}
by Lemma~\ref{lem:nested}. The inequality is an
equality whenever $R\le G_i$ almost surely. If $G_i=\max_j G_j>q$ then $R\le G_i$, because
$q<G_i$ and $Z_j\le G_j\le G_i$ for every $j$, so equality holds throughout for every
index-maximal arm. Hence $\mathcal V(S)=\max\{q,\max_i(\E[\mathcal V(S')]-c(s_i))\}$ at every state, attained by
stopping when $\max_i G_i\le q$ and by every index-maximal arm otherwise. In the second case
stopping does not attain it: for an index-maximal arm, $\E[(W_i-G_i)^+]=c(s_i)>0$ gives
$\Prob(W_i>G_i)>0$, and on that event $Z_i=G_i>q$, so $\mathcal V(S)>q$. This is the Bellman equation for
$V^\star+K$; the ladders are finite, $\mathcal V$ agrees with $V^\star+K$ at every state with no stage
left, and backward induction on the stages remaining gives $\mathcal V=V^\star+K$ everywhere. A policy that
at every state attains the maximum in the Bellman equation of a finite-horizon problem is optimal.
\end{proof}

At the boundary $\max_iG_i=q$, stopping is optimal and advancing any index-maximal arm is
also optimal by the same residual calculation. The stated policy chooses to stop on that
tie; optimal stopping is not asserted to be unique. At equal maximal indices above $q$, any
measurable tie-breaking rule is valid. No continuity of the reading distributions is needed.

Algorithm~\ref{alg:staged-index} combines the backward computation with the
policy of Theorem~\ref{thm:index}. The function returns an index and a conditional
covered-call law; its recursive calls integrate over hypothetical futures while
other arms' observed histories remain fixed. Only the policy's advance observes
a reading. Appendix~\ref{app:computation} gives finite-tree calculations,
complexity and the qualification for continuous laws.

\begin{algorithm}[htbp]
\caption{Staged index computation and search}
\label{alg:staged-index}
\small
\begin{algorithmic}[1]
\Require Initial state $S_0$, fixed planning law, ladders, collectable sets and costs
\Function{CoveredCall}{$i,T$}
  \If{arm $i$ has no collectable stage left at $T$}
    \State \Return $(-\infty,\delta_{-\infty})$ \Comment{point mass at $-\infty$}
  \EndIf
  \State For each possible next reading $y$, obtain
  \Statex \hspace{\algorithmicindent} $(\cdot,\mu_y)\gets\Call{CoveredCall}{i,F(T,i,y)}$
  \State Form the law of $W=\max\{\Yc(Y),Z_Y\}$ using
  \Statex \hspace{\algorithmicindent} $Y\sim p_\theta(\cdot\mid T,i)$ and $Z_Y\mid Y=y\sim\mu_y$
  \State Solve for the unique $g$ with $\E[(W-g)^+\mid T,i]=c(T,i)$
  \State \Return $(g,\operatorname{Law}(\min\{g,W\}\mid T,i))$
\EndFunction
\Procedure{StagedIndex}{$S_0$}
  \State $S\gets S_0$
  \While{$\cI(S)\ne\varnothing$}
    \ForAll{$i\in\cI(S)$}
      \State $(G_i,\cdot)\gets\Call{CoveredCall}{i,S}$
    \EndFor
    \If{$\max_{i\in\cI(S)}G_i\le q(S)$}
      \State \textbf{break}
    \EndIf
    \State $i\gets\min\argmax_{j\in\cI(S)}G_j$
    \State Pay $c(S,i)$ and observe the next reading $Y$
    \State $S\gets F(S,i,Y)$; update predictions using this observation
  \EndWhile
  \State \Return incumbent $q(S)$ and net utility $h(S)=q(S)-K(S)$
\EndProcedure
\end{algorithmic}
\end{algorithm}

The policy takes at most $L(S_0)$ advances. Under a coherent shared law, it
recomputes indices from the updated posterior marginals after each observation;
Proposition~\ref{prop:sharing-robustness} gives its guarantee under that
proposition's conditions.

\paragraph{Established special cases.}
The theorem permits dependent readings along an arm through its own history; it requires
other arms' laws to remain unchanged when that arm is observed.

For terminal-only collection the index is the generalized strike price of \citet[Appendix~G]{kleinberg2016descending}, and the recursion is that of \citet[Section~4, Definition~4]{bowers2024nested}. If every reading is collectable and its scalar transitions are first-order Markov with fixed positive stage costs, the policy is the multi-line
generalized reservation rule of \citet[Theorem~4.4]{yang2025markovian}. Independent readings
along a line also recover the corresponding order-constrained result of
\citet{boodaghians2023order}. Our history-state recursion and the value identity represent
designated collection and costs in the same construction. The superprocess interpretation
in Appendix~\ref{app:remarks} supplies the connection to the classical index theorem.

\begin{corollary}[Single-stage Pandora index]\label{cor:pbgi}
If $T_i=1$ and $\cK_i=\{1\}$ for every arm, $G_i$ solves
$\E[(Y_i-G_i)^+]=c_i$, and Theorem~\ref{thm:index} is Weitzman's theorem.
The policy is the independent-candidate PBGI of \citet{xie2024pbgi}.
\end{corollary}
\begin{proof}
With one collectable stage, $W(s)=\Yc=Y$ and $Z(s)=\min\{G(s),Y\}$, so the reservation
equation of Lemma~\ref{lem:nested} is $\E[(Y-G)^+]=c$ and the value identity of
Theorem~\ref{thm:index} is Weitzman's.
\end{proof}

Having only one collectable stage is not sufficient for this corollary if earlier
noncollectable inspection stages remain: those observations can support adaptive abandonment
and still require the nested recursion.

\subsection{Exact losses of arbitrary policies}
\label{sec:loss}
The covered call values do more than identify the optimal action: they price every suboptimal one
exactly, so the loss of any policy against the optimum can be located, not only bounded.

\begin{theorem}[Loss decomposition]\label{thm:loss}
Under the hypotheses of Theorem~\ref{thm:index}, at a state $S$ with incumbent $q$ let
\[
\Delta_{\rm stop}(S)=\E\bigl[(\max_jZ_j(s_j)-q)^+\bigr],
\]
and, for $i\in\cI(S)$ with $O_i=\max\{q,\max_{j\ne i}Z_j(s_j)\}$, which is independent of arm
$i$'s readings,
\[
\Delta_i(S)=c(s_i)-\E\bigl[(W_i(s_i)-\max\{O_i,G_i(s_i)\})^+\bigr],
\]
which equals $\E[\min\{(W_i-G_i)^+,(O_i-G_i)^+\}]$ when $G_i(s_i)>-\infty$. Then
$V^\star(S)-h(S)=\Delta_{\rm stop}(S)$, $V^\star(S)-Q^\star(S,i)=\Delta_i(S)\ge0$, and every
policy $\pi$, advancing arm $i_b$ at its $b$-th step, has
\[
V^\star(S)-V_\pi(S)=\E_{\theta,\pi}\Bigl[\sum_{b<\tau_\pi}\Delta_{i_b}(S_b)+\Delta_{\rm stop}(S_{\tau_\pi})\Bigm|S\Bigr].
\]
\end{theorem}
When $G_i=-\infty$, $W_i=-\infty$ and the first formula gives $\Delta_i=c(s_i)$, since
$O_i\ge q$ is finite. The expression involving $W_i-G_i$ is used only for finite indices.
\begin{proof}
With $\mathcal V$ as in the proof of Theorem~\ref{thm:index}, $V^\star+K=\mathcal V$ and $h=q-K$, so
$V^\star(S)-h(S)=\E[\max\{q,\max_jZ_j\}]-q=\Delta_{\rm stop}(S)$.
Condition on the next actual reading of $i$ and apply the tower property as in
Theorem~\ref{thm:index}; the added cost is $c(s_i)$. This gives
$Q^\star(S,i)+K+c(s_i)=\E[\max\{O_i,W_i\}]$ while
$V^\star(S)+K=\E[\max\{O_i,\min\{G_i,W_i\}\}]$, and
$\max\{O,W\}-\max\{O,\min\{G,W\}\}=(W-\max\{O,G\})^+$ gives $V^\star(S)-Q^\star(S,i)=\Delta_i(S)$,
nonnegative by the Bellman equation. The advance reveals only the next reading,
not the full-path variable $W_i$. When $G_i$ is finite, $c(s_i)=\E[(W_i-G_i)^+]$ by
Lemma~\ref{lem:nested}, and $(W-G)^+-(W-\max\{O,G\})^+=\min\{(W-G)^+,(O-G)^+\}$ pointwise.
The policy identity is the performance-difference telescoping of these one-step losses along
$\pi$, given in Appendix~\ref{app:loss-proof}.
\end{proof}

The theorem is the performance-difference identity of dynamic programming with both of its terms
exact. For finite $G_i$, $\Delta_i=0$ exactly when $O_i\le G_i$ almost surely:
the reservation equation and positive cost give $\Prob(W_i>G_i)>0$, and $W_i$ is
independent of $O_i$. Thus an advance costs nothing against the optimum exactly
when the rest of the pool, incumbent included, almost surely cannot overtake its
index; otherwise it costs the expected
overshoot of the arm's nested value above its index, capped by how far the rest of the pool can
exceed that index. A stop costs the value of the covered call values above the incumbent. Every term is a functional of the arms' covered call distributions. The identity applies to any policy, including one that advances through states at which the index policy would stop or switch. Evaluating the distributions makes each such departure measurable against the same optimum.

\section{Comparative Statics: Rights, Costs and Resumption}
\label{sec:comparative}

The covered call values also price changes in the problem itself. In this section consider two per-arm problems $\mathcal P$ and $\mathcal P'$, each satisfying Assumption~\ref{assm:standing}, with the same arm path laws, so that each arm's readings are the same random variables under both. Say that $\mathcal P'$ \emph{dominates} $\mathcal P$ if $\cK_i\subseteq\cK_i'$ and
$c_i'(s)\le c_i(s)$ for every arm and own state: more readings may be kept and
no advance costs more. Corresponding states share the observed context, hence the own states $s_i$, and have incumbents $q\le q'$; the state of $\mathcal P'$ corresponding to $S$ is written $\bar S$. Primes on $W$, $Z$, $G$, $\Phi$, $\Yc$, $c$, $q$, $V$ and $U$ denote quantities of $\mathcal P'$ throughout this section, while $s'$ keeps its meaning as the successor own state.

\begin{lemma}[Pathwise monotonicity]\label{lem:pathwise}
If $\mathcal P'$ dominates $\mathcal P$, then at every own state $s$,
$W'(s)\ge W(s)$ and $Z'(s)\ge Z(s)$ on every remaining path, and $G'(s)\ge G(s)$.
\end{lemma}
\begin{proof}
Backward induction on $\nu(s)$; all three are $-\infty$ when $\nu(s)=0$. At $s$
with $\nu(s)\ge1$, the collectable reading satisfies $\Yc'\ge\Yc$ because
$\cK\subseteq\cK'$, and $Z'(s')\ge Z(s')$ by the hypothesis at $s'$, so
$W'(s)=\max\{\Yc',Z'(s')\}\ge W(s)$. By Lemma~\ref{lem:nested},
$\Phi'(s;r)=\E[(W'(s)-r)^+]-c'(s)\ge\E[(W(s)-r)^+]-c(s)=\Phi(s;r)$ for every
$r$, so the positivity set of $\Phi'$ contains that of $\Phi$ and
Proposition~\ref{prop:index} gives $G'(s)\ge G(s)$. Then
$Z'(s)=\min\{G'(s),W'(s)\}\ge\min\{G(s),W(s)\}=Z(s)$.
\end{proof}

\begin{proposition}[Exact comparative statics]\label{prop:comparative}
Let $\mathcal P'$ dominate $\mathcal P$. At corresponding states $S$ and $\bar S$, with $U=V^\star+K$ as in Section~\ref{sec:setting}, \[ U'(\bar S)-U(S)=\E\Bigl[\bigl(\max\{q',\max_iZ_i'(s_i)\}-\max\{q,\max_iZ_i(s_i)\}\bigr)^+\Bigr]\ge0,
\]
the expectation coupling $Z_i$ and $Z_i'$ on arm $i$'s remaining path under
$\mathsf P_S$. At a common initial state the value of the additional rights or
lower costs is $\E[(\max_iZ_i'-\max\{q,\max_iZ_i\})^+]$, which vanishes if and
only if $\max_iZ_i'\le\max\{q,\max_iZ_i\}$ almost surely.
\end{proposition}
\begin{proof}
Theorem~\ref{thm:index} applies to both problems on the common product space, giving $U'(\bar S)-U(S)=\E[\max\{q',\max_iZ_i'\}-\max\{q,\max_iZ_i\}]$. By
Lemma~\ref{lem:pathwise} and $q'\ge q$ the integrand is nonnegative on every
path, so it equals its positive part.
\end{proof}

The identity follows from pathwise monotonicity and the value representation. Its use is
to price additional collection rights or lower costs from coupled arm-wise covered-call
values, without solving either joint Bellman problem. It covers enlarging a collectable
set, removing a history-dependent surcharge, and lowering any own-history cost.
For one-shot boxes, adding a box or lowering a cost raises the stopping frontier
\citep[Proposition~2]{bao2026interdependent}. The identity does not cover changes to the
readings' law, such as adding an observation stage or changing signal accuracy.
The next result prices the opposite error: planning
as if fewer readings could be kept.

\begin{proposition}[Planning with a smaller collectable set]\label{prop:fewer-rights}
Let $\mathcal P'$ dominate $\mathcal P$ with the same costs, so that only the collectable sets differ. Let $\pi_{\cK}$ be the policy for $\mathcal P'$ that stops when $\max_iG_i(s_i)\le q'$ and otherwise advances an arm maximizing the $\mathcal P$-index $G_i(s_i)$, where $q'$ is the incumbent of $\mathcal P'$. Then at corresponding states \[ V^\star(S)\le V_{\pi_{\cK}}(\bar S)\le V^{\star\prime}(\bar S). \] \end{proposition} Its loss against the optimum of $\mathcal P'$ is therefore at most the value of the additional rights; both ends are attained in the study of Appendix~\ref{app:comparative-results}.

Appendix~\ref{app:comparative} proves this and the next result. Finally, the
loss decomposition prices the freedom to return to a paused arm. Fix a starting state
$S_0$. A policy \emph{resumes} if it returns to an arm that it has advanced since $S_0$
after advancing a different arm; a \emph{no-return} policy never resumes, and $V_{\rm NR}(S_0)$
is its optimal value. This differs from search without recall, or committed search
\citep[Section~5]{gupta2019markovian}, which forbids returning to an earlier prize:
a no-return policy retains every collectable reading.

On a path where the index resumes, a no-return policy must depart from its
trajectory by that advance. We lower-bound the resulting loss by the smallest
expected Bellman penalty at this first departure; later losses are nonnegative.

To preserve return status in the bound, augment a descendant history with
$\widehat S=(S,\ell,D)$, where $\ell$ is the last arm advanced since $S_0$
($\ell=0$ initially) and $D$ is the set of arms advanced and subsequently left since
$S_0$. An advance of $i$ sets $\ell^+=i$ and
$D^+=D\cup\{\ell\}$ if $\ell\notin\{0,i\}$, and $D^+=D$ otherwise.
It is a resumption exactly when $i\in D$ and $i\ne\ell$. This bookkeeping is retained along the
whole run, including when the last arm is exhausted.
For the index policy with its fixed tie order, let $\delta(S)$ be the smallest
loss from Theorem~\ref{thm:loss} of an action other than $\pi_{\rm IX}(S)$,
with $\min\varnothing=+\infty$. Define $\widehat\rho(\widehat S)$ by finite backward
recursion: it is zero if the index policy has zero probability of a future resumption
from this augmented history, equals $\delta(S)$ if its next action resumes, and otherwise
\[
 \widehat\rho(\widehat S)
 =\min\{\delta(S),\E[\widehat\rho(\widehat S_1)\mid\widehat S]\},
 \qquad \rho(S_0)=\widehat\rho(S_0,0,\varnothing),
\]
where $\widehat S_1$ is the successor under the index policy with the bookkeeping above.
The minimum compares departing now, at penalty $\delta(S)$, with following the
index one more step and facing a later departure. At the first resumption,
departure is compulsory. This first-departure calculation gives a lower bound
on the total resumption gap.

\begin{proposition}[Value of resumption]\label{prop:resumption}
Under the hypotheses of Theorem~\ref{thm:index}:
\begin{enumerate}
\item[(i)] $V^\star(S_0)-V_{\rm NR}(S_0)\ge\rho(S_0)\ge0$, and $V_{\rm NR}(S_0)=V^\star(S_0)$ when $\pi_{\rm IX}$ almost surely never resumes from $S_0$.
\item[(ii)] If at every state at which $\pi_{\rm IX}$ advances from $S_0$, before or at its first resumption, the index-maximal arm is unique, then $\rho(S_0)>0$ exactly when $\pi_{\rm IX}$ resumes from $S_0$
 with positive probability.
\item[(iii)] If every reading is a deterministic function of the arm's own
 state, $\pi_{\rm IX}$ never resumes under any tie order, so $V_{\rm NR}=V^\star$.
\end{enumerate}
\end{proposition}

With deterministic readings an advance either raises the arm's index by its cost
or lifts the incumbent above its previous index, which rules out resumption.
Stochastic observations can make resumption valuable by resolving uncertainty in
collectable rewards, even without changing any unobserved reward's law.
Example~\ref{ex:resumption-reward} in Appendix~\ref{app:comparative} gives a
two-arm construction and calculates the values with and without resumption.

Section~\ref{sec:validation} and Appendix~\ref{app:comparative-results}
evaluate all three results exactly on the staging grid.

\section{Information Sharing Across Arms}
\label{sec:robust}

Sections~\ref{sec:main} and~\ref{sec:comparative} assumed a per-arm planning law. Under the general law of Assumption~\ref{assm:standing}, define
$G_i(S)=G_i^{(\nu_i)}(S)$ by advancing only $i$ from the current history; other arms'
observations stay fixed along this hypothetical run, so the local recursion still applies,
and the index policy recomputes these indices after each actual observation. Shared
observations can change untouched arms' laws; freezing hyperparameters does not restore
independence.

\begin{proposition}[No universally optimal own-arm-law policy]\label{prop:no-local}
There are two instances with the same incumbent, costs, available stages and separate laws of
each arm's remaining readings, one independent and one shared, with different uniquely optimal
first actions. Consequently no single rule whose inputs consist only of those quantities is
optimal for all joint planning laws, including randomized rules.
\end{proposition}

The construction couples a noncollectable signal in one arm to another arm's collectable
reward while preserving their separate laws; for one-shot boxes,
\citet{bao2026interdependent} show similarly that fixed box indices cannot be optimal under
interdependence. A positive standalone run value still certifies that stopping is suboptimal,
and under a forward-closure condition the index stopping region is exactly optimal
(Proposition~\ref{prop:stop-any}, Appendix~\ref{app:sharing}).

\paragraph{Collection reshapes acquisition under a shared state.}
Sharing also interacts with designated collection: the same joint law of readings can
yield different acquisition orders under different collection rights.

\begin{proposition}[Collection-induced acquisition switching]\label{prop:switching}
There is a two-arm family with a common binary latent state of prior $p$, a nondegenerate
closed interval of terminal costs, and thresholds $0<\tau_1<\tau_2(c)<\tau_3(c)<1$ depending
continuously on the cost $c$ in that interval, with the following property. If arm $B$'s
intermediate reading is collectable, whether or not $A$'s is, the uniquely optimal first
action advances $A$ for $p\in(0,\tau_1)\cup(\tau_2(c),\tau_3(c))$ and advances $B$ for
$p\in(\tau_1,\tau_2(c))\cup(\tau_3(c),1)$. If it is not collectable, whether or not $A$'s is,
advancing $A$ is the uniquely optimal first action for every $p\in(0,1)$.
\end{proposition}

In the construction, arm $A$ has the higher terminal success probability in one
latent state, while $B$ offers the larger usable intermediate score. Keeping that score adds an
option to avoid a terminal inspection; its value competes with terminal reward
and information through the subsequent decisions. The alternating regions show
that first inspection need not vary monotonically with the prior.

Only the collection rights differ across the four cases, so the set of priors at which $A$
is inspected first is disconnected exactly when $B$'s intermediate score can be kept.
A prior-dependent change of first inspection already occurs in
one-shot interdependent search \citep[Example~1]{bao2026interdependent}. The family,
its thresholds, a fixed-cost first-action table and the proof are in
Appendix~\ref{app:switching}; the proof is an exact
computer-assisted Bellman certificate over the field of rational functions of $c$, uniform
over the cost interval.

\paragraph{Robustness under a coherent shared law.}
A quantitative guarantee follows by specializing approximate-planning bounds
\citep[Theorem~9 and Remark~11]{subramanian2022ais} to the independent value representation.
Restrict further: all readings are coordinates of complete arm paths with one fixed joint law
$\mathsf P$, predictions are its conditional laws, and each arm's next cost depends only on
its own observed history. Dependence within and across paths is arbitrary; cross-arm cost
changes are excluded. Let $\mathsf P_{j,S}$ be the posterior marginal law of arm $j$'s
remaining path at $S$, $Z_j^S$ its covered call under that law, and
$v_{j,S}(r)=\E_{\mathsf P_{j,S}}\max\{r,Z_j^S\}$ its local keep-or-run value. For the actual
successor $S'=F(S,i,Y)$, the discrepancy
\[
 \varepsilon_i(S)=\E\!\left[
   \sum_{j\ne i}\sup_{r\in[q(S'),y_{\max}]}
       |v_{j,S'}(r)-v_{j,S}(r)|\,\middle|\,S,i\right]
\]
measures how much an observation changes untouched arms' local values; each supremum is the
stop-loss metric between the two capped variables clipped at the new incumbent
\citep[Section~5, Example~C]{muller1997ipm}.

\begin{proposition}[Robustness to bounded sharing]\label{prop:sharing-robustness}
Under Assumption~\ref{assm:standing} and the coherent-law/local-cost restriction
above, let $H=L(S_0)$. If $\varepsilon_i(S)\le\bar\varepsilon$ almost surely
along every admissible policy from $S_0$, then the policy $\pi_{\rm IX}$,
recomputed from the true posterior marginal laws after every observation, satisfies
\[
 0\le V^\star(S_0)-V_{\pi_{\rm IX}}(S_0)\le 2H\bar\varepsilon.
\]
\end{proposition}

The bound covers premature stops and incorrect advances and needs no separation between
indices; under independence $\varepsilon_i=0$. The condition concerns every future feasible
history, including states at which the index would already have stopped, so root covariance
alone does not suffice, and it does not control posterior misspecification or numerical index
error. Appendix~\ref{app:sharing-robustness} proves the bound through an explicit signed
sharing correction and gives a path-total-variation envelope.

\paragraph{Sharing through a learned common component.}
In Bayesian optimization, sharing typically flows through a component common to all
candidates, such as a task-level parameter of a learning-curve model learned from previous
tasks and from the current run. Suppose arm paths are conditionally independent given a latent
$\Theta$ on a standard Borel space, with posterior $\mu_S$ at state $S$, so that
$\mathsf P_{j,S}=\int\mathsf P_j^\vartheta(\cdot\mid s_j)\,\mu_S(d\vartheta)$ for
measurable component path kernels $\mathsf P_j^\vartheta$. Let
\[
 \Lambda_j(S)=\sup_{\vartheta,\vartheta'}\operatorname{TV}\bigl(\mathsf P_j^\vartheta(\cdot\mid s_j),\mathsf P_j^{\vartheta'}(\cdot\mid s_j)\bigr),\qquad
 \eta_i(S)=\E_{\vartheta\sim\mu_S}\operatorname{TV}\bigl(\mathsf P_i^\vartheta(Y\in\cdot\mid s_i),\mathsf P_{i,S}(Y\in\cdot)\bigr),
\]
be the \emph{sensitivity} of arm $j$'s remaining path to the component and the posterior
\emph{dispersion} of arm $i$'s next reading, with
$\operatorname{TV}(\mu,\nu)=\sup_A|\mu(A)-\nu(A)|$, and let
$B_j(S)=y_{\max}-y_{\min}+\nu_j(S)c_{\max}$ if arm $j$ has a collectable stage left and
$B_j(S)=0$ otherwise. Assume the sensitivities $\Lambda_j(S)$ are measurable;
this is automatic for a finite component space.

\begin{corollary}[Learned common component]\label{cor:component}
Under Assumption~\ref{assm:standing} and the coherent-law and arm-local-cost restriction of Proposition~\ref{prop:sharing-robustness}, with arm paths conditionally independent given $\Theta$,
\[
 \varepsilon_i(S)\le\eta_i(S)\sum_{j\ne i}B_j(S)\Lambda_j(S),
 \qquad\text{so}\qquad
 0\le V^\star(S_0)-V_{\pi_{\rm IX}}(S_0)\le 2H\sup_{S,i}\eta_i(S)\sum_{j\ne i}B_j(S)\Lambda_j(S)
\]
over feasible states and advances. Moreover
$\eta_i(S)\le\E_{\vartheta,\vartheta'\sim\mu_S\otimes\mu_S}
\operatorname{TV}(\mathsf P_i^\vartheta(Y\in\cdot\mid s_i),\mathsf P_i^{\vartheta'}(Y\in\cdot\mid s_i))$,
the expected disagreement of two posterior draws. For binary $\Theta$ with
$p=\mu_S(\Theta=1)$, $\eta_i(S)=2p(1-p)\operatorname{TV}(\mathsf P_i^0(Y\in\cdot\mid s_i),\mathsf P_i^1(Y\in\cdot\mid s_i))$
and $\Lambda_j(S)$ is the separation of the two component path laws.
\end{corollary}

Sensitivity measures how far a candidate's law can move with the component, dispersion how
much one more reading can move the belief about it. For fixed horizon, reading and cost bounds,
a vanishing uniform envelope on the right-hand side gives a vanishing regret bound.
For a finite component space, posterior mass tending to one on one component makes
the local dispersion vanish. With continuous components, posterior concentration needs
additional regularity, such as total-variation continuity of the next-reading kernels;
weak concentration alone does not suffice. In either case the regret guarantee still
requires control over all feasible future histories, including rare ones whose beliefs
may be less concentrated. Appendix~\ref{app:sharing-robustness} gives a noiseless
counterexample to automatic dispersion decay.
Mixtures of product laws are also the structure under which correlated
one-shot search admits efficient approximation \citep{chawla2023adaptivecorrelations}.
Appendix~\ref{app:sharing-robustness} gives the proof; Section~\ref{sec:validation} and
Appendix~\ref{app:sharing-results} report checks of the signed correction and bounds;
Appendix~\ref{app:sharing-controls} contrasts a positive-regret example with a
shared nuisance whose local-value envelope vanishes, and reports a sweep in which
only the prior on a binary component changes and the uniform envelopes decrease.

\section{Numerical Verification}
\label{sec:validation}

Exact finite-law checks compare the representation and losses with direct joint
Bellman optimization, using rational arithmetic throughout
(Table~\ref{tab:verification-summary}). On independent binary-tree instances,
the optimal value, maximal-index actions and every stop/advance loss agree.
On the staging grid, each diagnostic policy's utility loss equals its accumulated
action losses. Across dominated problem pairs, value differences equal the
covered-call overshoot; both ends of the smaller-collectable-set bound are
attained. A separate no-return recursion confirms that $\rho$ never exceeds the
resumption gap, including cases of equality (Table~\ref{tab:comparative}).

Under shared laws, the signed residual, policy telescoping and envelope bounds
hold at every feasible prefix, including histories the index never visits.
A constructed example has positive index regret, while a shared nuisance has
zero local-value envelope despite nonzero path-TV shifts
(Table~\ref{tab:robustness-bounds}). Across the tested priors in a binary-component
sweep, the largest uniform call and path-TV bounds decrease as the prior moves
toward zero, and regret is zero at the four tested priors below $1/2$
(Table~\ref{tab:concentration}). This illustrates the uniform-envelope condition
in Corollary~\ref{cor:component} for these fixed finite models. These checks
directly support the sharing guarantee and its value-sensitive discrepancy.
Appendix~\ref{app:experiments}
places each study's law beside its checks and results;
Algorithm~\ref{alg:staged-index} gives the policy and backward index computation;
Appendix~\ref{app:computation} gives finite-tree implementation details and complexity.
The checks supplement the proofs under known laws; they establish neither
learned-surrogate accuracy nor practical optimization performance.

\section{Conclusion}
\label{sec:limitations}

The covered-call representation unifies designated collection, history-dependent prediction
and own-state costs. It gives the independent value, optimal policy and exact action losses,
recovers established special cases, prices collection rights and costs, bounds the value of resumption, and
quantifies how posterior updates under sharing perturb those guarantees through
changes in local continuation values. A common-component envelope relates those
changes to predictive total-variation dispersion; its decay must be controlled
over all feasible future histories.

Exact optimality requires finite independent arm processes, arm-local availability and
costs, and readings that follow the planning law; general information sharing invalidates the
proof, and the forward-closure stopping condition is not established for shared surrogates.
The sharing bound requires a coherent joint path law, arm-local costs, true posterior
marginals and an envelope over every future feasible history; it is conservative in the exact
examples and certifies neither posterior accuracy nor numerical index error.
Weak posterior concentration alone does not imply that this envelope vanishes.
An unbounded Gaussian predictive law also requires an extension of the stated
bounded-law assumptions. Continuous
reading laws require integration or discretization whose policy error is not bounded here,
observation trees are exponential in depth, and the objective is the best observed
collectable score less cost, not latent performance under noisy readings. The studies use
two arms, few outcomes, short ladders and known laws; they measure neither learned-surrogate
accuracy nor robustness to a misspecified law and establish no empirical superiority for
hyperparameter optimization. Scaling, misspecification, continuous candidate optimization and
comparisons with published methods on their own surrogates remain open for the staged
index; \citet{vu2026coststop} report such comparisons for the one-step rule.

\bibliographystyle{plainnat}
\bibliography{p2_arxiv}
\appendix
\section{Notation and Local Stopping Details}
\label{app:proof-details}

\subsection{Notation and literature conventions}
\label{app:notation}

An arm's own state $s_i$ consists of its stage and observation history;
$S$ records the full search history. The remaining depth is $\nu_i=T_i-t_i$,
$\cK_i$ is the collectable set, and $K(S)$ is sunk cost. The forced-run gain
$\Phi(s;r)$ excludes immediate stopping; the keep-or-run value is
$r+\Phi(s;r)^+$. At exhaustion that value remains $r$. A suffix without a
collectable stage has index and cap $-\infty$; differences such as $W-G$ are
used only for finite $G$. The full-path variables $W,Z$ are not disclosed by
one advance.

For terminal-only collection, $G_i,Z_i$ correspond to the nested strike and
capped values $\sigma_i^{(\ell)},\kappa_i^{(\ell)}$ of
\citet[Definition~4]{bowers2024nested}, with $\ell=t_i+1$, and $W_i$ to their
next capped value. Our forced gain differs from the total equivalent-reward
function called $\Phi$ by \citet[Definition~3.2]{yang2025markovian}.

The per-arm path law is $\mathsf P_i^{s_i}$. Under sharing, $\mathsf P$ is the
coherent joint law, $\mathsf P_{j,S}$ a posterior arm marginal, and
$\mathsf Q_S=\bigotimes_j\mathsf P_{j,S}$ the synthetic product reference.
The action value $Q_\pi(S,i)=\E[V_\pi(S')]$ includes the advance cost through
$K(S')$; $U_\pi=V_\pi+K$ removes sunk costs. Expectations at a starting state
use its specified kernel and count further advances from zero, including for
exact real-valued histories of probability zero. Conditional on that state,
all remaining rewards and costs are bounded.

\subsection{Stopping-rule attainment}

\begin{lemma}[Finite-horizon arm-run recursion]\label{lem:run-dp}
Under Assumption~\ref{assm:standing}, the forced-run value
$(S,r)\mapsto\Phi_i^{(k)}(S;r)$ is jointly Borel measurable on states at which $i$ is
available. Its supremum is attained by a stopping rule chosen measurably in $(S,r)$.
This holds for an arbitrary fixed planning kernel, including a shared one.
\end{lemma}
\begin{proof}
For an auxiliary outside value $r$, set $U_{i,0}(S;r)=0$. For $k\ge1$ and an available $i$,
define backwards
\begin{align*}
B_{i,k}(S;r)&=-c(S,i)+\int\bigl[(\tilde y-r)^+
 +U_{i,k-1}(F(S,i,y);\max\{r,\tilde y\})\bigr]\,p_\theta(dy\mid S,i),\\
U_{i,k}(S;r)&=\max\{0,B_{i,k}(S;r)\},
\end{align*}
where $\tilde y=y$ if this advance is collectable and $-\infty$ otherwise. Put $U_{i,k}=0$
at exhausted arms. All functions are jointly Borel by induction: the successor, cost and kernel
are measurable, integration preserves measurability, and the action maximum is over two
choices. For each finite $r$ the integrands and values are bounded by the reading bounds and
the finite remaining cost bound.

Let $m$ readings have been seen along a run and let $r_m=\max\{r,M_m\}$. For any additional
future maximum $M'$, the identity
\[
(\max\{M_m,M'\}-r)^+-(M_m-r)^+=(M'-r_m)^+
\]
shows that the earned improvement can be separated from future improvement. The full planning
state in the recursion is still the actual $S^{(m)}$; replacing $r$ by $r_m$ does not replace
its incumbent, refit its law, or change its costs. Thus $B_{i,k}$ is the value of forcing one
advance followed by an optimal continue-or-stop decision, and $U_{i,k}$ includes stopping
immediately. Backward induction gives
$B_{i,k}(S;r)=\Phi_i^{(k)}(S;r)$ and $U_{i,k}(S;r)=\Phi_i^{(k)}(S;r)^+$ whenever $i$ is
available. After the forced first advance, continue precisely where the remaining $B$ is
positive and stop on a tie or at the cap. These are measurable decisions and give a stopping
time bounded by $\min\{k,\nu_i\}$ attaining the stated value.
\end{proof}

For a per-arm law, the recursion depends on $S$ only through $s_i$ and the auxiliary value.
This justifies the local Bellman equation in Section~\ref{sec:nested} without assuming that
the latest scalar reading is Markov. Measurability of the reservation indices follows from
their rational-threshold representation in Section~\ref{sec:setting}; no selection over
uncountably many stopping rules or candidate actions is needed.

\begin{proof}[Proof of Proposition~\ref{prop:index}]
The finite-horizon recursion in Lemma~\ref{lem:run-dp} gives a jointly measurable value and
an attaining stopping rule. In particular, the supremum here does not require an unproved
measurable selection from a family of stopping times.
For a fixed run, its positive-part payoff is nonincreasing and $1$-Lipschitz in $r$; these
properties pass to expectation and supremum. The value is finite, bounded above by
$(y_{\max}-r)^+-c(S,i)$ and below by the one-advance run. Without a collectable stage within
the cap every run has strictly negative value, so $G=-\infty$. Otherwise the committed run to
the next collectable stage has positive value for $r<y_{\min}-\mu_i c_{\max}$, while the upper
bound is nonpositive for $r\ge y_{\max}-c(S,i)$. The positivity set is therefore a nonempty
lower interval. Continuity excludes a positive endpoint and forces the value there to be zero,
giving the strict threshold statement.
Increasing the cap enlarges the set of runs and cannot decrease the index.
\end{proof}

\subsection{Proof of the policy identity in Theorem~\ref{thm:loss}}
\label{app:loss-proof}

\begin{proof}
Put $H=L(S)$ and $\tau=\tau_\pi\le H$. Let $\mathcal F_b$ contain the observed
history and any independent policy randomness used to choose the next action. On $b<\tau$,
the planning kernel gives
\[
\E[V^\star(S_{b+1})\mid\mathcal F_b]=Q^\star(S_b,i_b)
=V^\star(S_b)-\Delta_{i_b}(S_b).
\]
The event $\{b<\tau\}$ is $\mathcal F_b$-measurable. Freeze the process after it stops and
sum this identity over the deterministic range $b=0,\ldots,H-1$, with indicators
$\mathbf1\{b<\tau\}$. The tower property and telescoping give
\[
V^\star(S)-\E[V^\star(S_\tau)]
=\E\!\left[\sum_{b<\tau}\Delta_{i_b}(S_b)\right].
\]
Finally $V^\star(S_\tau)=h(S_\tau)+\Delta_{\rm stop}(S_\tau)$ and
$V_\pi(S)=\E[h(S_\tau)]$. Every sum is finite and every term bounded conditional on the
starting state, so no optional-stopping or interchange-of-infinite-limits assumption is used.
\end{proof}

\subsection{Continuous intermediate collection}

\begin{example}[Continuous, dependent readings with intermediate collection]
\label{ex:continuous-collection}
Consider one arm with two collectable stages, $Y_1=U$ and $Y_2=1-U$ for
$U\sim\operatorname{Uniform}[0,1]$, and costs $c_1=1/8$, $c_2=1/4$.
The second reading is deterministic conditional on the first, so the within-arm dependence
is perfect and neither stage is an independent box. At the state after $U=u$,
\[
G_2(u)=Z_2(u)=\tfrac34-u,\qquad
W_1=\max\{U,\tfrac34-U\}.
\]
For $r\in[3/8,3/4]$, integration over the two disjoint exceedance intervals gives
\[
\E[(W_1-r)^+]=\tfrac12(1-r)^2+\tfrac12(\tfrac34-r)^2.
\]
The unique reservation root is therefore
$G_1=(7-\sqrt7)/8$, which lies in this interval. At incumbent $q=1/2$ and cumulative cost $K=0$,
\[
\Phi_1(1/2)=\tfrac{5}{32}-\tfrac18=\tfrac1{32},\qquad
V^\star=\tfrac{17}{32}.
\]
After the first reading, the index policy continues to the second exactly when
$\tfrac34-u>\max\{1/2,u\}$, or $u<1/4$; equality may be stopped. Direct integration of that
policy's reward minus both incurred costs gives the same value $17/32$. This is an analytic
illustration of the recursion with a continuous marginal law and an intermediate collectable
reading, not a finite-quadrature optimality claim.
\end{example}

\section{Connection to Classical Superprocess Indices}
\label{app:remarks}

\begin{remark}[The theorem as a superprocess index theorem]\label{rem:whittle}
Adding to an arm a free \emph{keep} transition that pays the running maximum of its collectable
readings would encode the collectable set in the Markovian boxes of \citet{gupta2019markovian},
but it gives the arm two actions and leaves the class of Markov chains for that of decision
processes, which is why the reduction is to the superprocess theorem and not to the Markovian box.
In the language of \citet{whittle1980gittins} and \citet[\S5.2.2]{scully2025gittins}, who call the setting MDP selection, a
nonterminal arm at state $s$ with best previously collectable reading $M$ can \emph{advance}
at reward $-c(s)$ or \emph{keep} $M$ and terminate the selection. When none has been observed,
$M=-\infty$ and keeping is unavailable. Including an outside retirement value $\alpha$ gives
the local value
\[
r+\Phi(s;r)^+,\qquad r=\max\{\alpha,M\},
\]
by Lemma~\ref{lem:nested}. Here advance is compared with the better of \emph{both} immediate
options, not with keeping $M$ alone: retirement can exceed $M$.

Whittle's condition asks for one non-retirement action, chosen independently of $\alpha$,
such that this action or retirement is optimal for every $\alpha$. The choice is
\begin{center}
\begin{tabular}{lll}
\toprule
Condition at the arm state & Retained action & Retirement optimal for \\
\midrule
$M> -\infty$, $M\ge G(s)$ & keep & $\alpha\ge M$ \\
$M<G(s)$ & advance & $\alpha\ge G(s)$ \\
\bottomrule
\end{tabular}
\end{center}
Below the displayed retirement threshold, the retained action is optimal;
at the threshold both may be chosen. Its identity depends only on the fixed
arm state, through $M$ and $G(s)$, while the choice to retire may depend on $\alpha$.
If both $M$ and $G(s)$ are $-\infty$, retirement is optimal for every finite $\alpha$;
any available advance may be the designated action in that vacuous case.
The resulting outside-option threshold is $\max\{M,G(s)\}$. With the incumbent represented
as a deterministic retirement option (equivalently, a dummy arm), it yields
$\pi_{\rm IX}$ in the terminating selection setting of \citet{glazebrook1979stoppable}. This is a connection to classical theory; the direct Bellman
verification in Theorem~\ref{thm:index} already establishes optimality for the stated model.
The nonobligatory-inspection hardness does not apply because an unobserved reward cannot be kept.
\end{remark}

\section{Proofs for Section~\ref{sec:comparative}}
\label{app:comparative}

\begin{proof}[Proof of Proposition~\ref{prop:fewer-rights}]
The upper bound holds because $\pi_{\cK}$ is admissible in $\mathcal P'$. For
the lower bound, induct on $L(S)$ over corresponding states, which share
the own states and sunk cost $K$ and have $q\le q'$. If $\pi_{\cK}$ stops at $\bar S$ then $\max_iG_i(s_i)\le q'$, and since $Z_i\le G_i$ pathwise,
Theorem~\ref{thm:index} gives
$V^\star(S)+K=\E[\max\{q,\max_iZ_i\}]\le\max\{q,\max_iG_i\}\le q'$, so
$V^\star(S)\le q'-K=V_{\pi_{\cK}}(\bar S)$. Otherwise $\pi_{\cK}$ advances an arm
$i$ with $G_i(s_i)=\max_jG_j(s_j)>q'\ge q$, which is an optimal action in
$\mathcal P$ at $S$ by Theorem~\ref{thm:index}. Both problems pay $c(s_i)$ and
reveal the same reading $Y$; the successors correspond, with incumbents
$\max\{q,\Yc\}\le\max\{q',\Yc'\}$ since $\Yc\le\Yc'$. Hence
$V^\star(S)=\E[V^\star(S_1)]\le\E[V_{\pi_{\cK}}(\bar S_1)]
=V_{\pi_{\cK}}(\bar S)$ by the induction hypothesis applied under the integral.
The successor values already include the advance cost through
$K(S_1)=K(\bar S_1)=K+c(s_i)$.
Appendix~\ref{app:comparative-results} exhibits cells attaining each bound.
\end{proof}

\begin{proof}[Proof of Proposition~\ref{prop:resumption}]
(i) The index policy is no-return whenever it never resumes, which gives the
equality case. Let $\pi$ be any no-return policy; conditional on its auxiliary randomness it is a deterministic no-return policy, so we may take it deterministic. Let $\sigma$ be the first
step at which $\pi$'s action differs from $\pi_{\rm IX}$'s at the state
reached, with $\sigma=\infty$ if they never differ. Up to $\sigma$ both
policies take the same actions and observe the same readings, so the law of
$(\widehat S_0,\ldots,\widehat S_\sigma)$ under $\pi$ is the law of the same stopped segment of
the $\pi_{\rm IX}$ trajectory, including the last-arm and left-arm bookkeeping
from the fixed root $S_0$. Since $\pi$ cannot resume, $\sigma$ is at most
the first time $\pi_{\rm IX}$ resumes. Theorem~\ref{thm:loss} writes
$V^\star(S_0)-V_\pi(S_0)$ as the expected sum of nonnegative losses along $\pi$,
one of which, on $\{\sigma<\infty\}$, is the loss of $\pi$'s action at
$S_\sigma$; that action differs from $\pi_{\rm IX}(S_\sigma)$, so its loss is at
least $\delta(S_\sigma)$. Hence
$V^\star(S_0)-V_\pi(S_0)\ge\E_{\pi_{\rm IX}}[\delta(S_\sigma)\mathbf 1\{\sigma<\infty\}]$.
The right side is the cost of a stopping time on the $\pi_{\rm IX}$ trajectory
that must stop no later than the first resumption and may run forever only on
branches without resumption. The infimum of this cost over such stopping times
satisfies the finite-horizon recursion defining $\widehat\rho$: it is zero where no
resumption can occur, equals $\delta$ at a resumption state, and otherwise is
the smaller of stopping now and continuing. Thus
$V^\star(S_0)-V_\pi(S_0)\ge\widehat\rho(S_0,0,\varnothing)=\rho(S_0)$
for every no-return $\pi$.

(ii) At a state $S$ where the index-maximal arm $i^\star$ is unique with
$G_{i^\star}>q$, every other action has positive loss. By
Lemma~\ref{lem:nested}, $\E[(W_{i^\star}-G_{i^\star})^+]=c(s_{i^\star})>0$,
so $W_{i^\star}>G_{i^\star}$ and hence $Z_{i^\star}=G_{i^\star}$ with positive
probability. Then $\Delta_{\rm stop}(S)=\E[(\max_jZ_j-q)^+]>0$. For an arm
$j\ne i^\star$ with finite index, $O_j\ge Z_{i^\star}=G_{i^\star}>G_j$ with
positive probability, so $\Prob(O_j>G_j)>0$ and
$\Delta_j(S)>0$ by the paragraph following Theorem~\ref{thm:loss}; if $G_j=-\infty$
then $\Delta_j(S)=c(s_j)>0$. Hence $\delta(S)>0$ at every state in the
hypothesis. The augmented recursion then gives $\widehat\rho(\widehat S)>0$
if and only if resumption from that augmented history has positive probability:
at a resumption state $\widehat\rho=\delta>0$, on branches without resumption
$\widehat\rho=0$, and elsewhere
$\widehat\rho=\min\{\delta,\E[\widehat\rho(\widehat S_1)]\}$ is positive exactly when
$\E[\widehat\rho(\widehat S_1)]>0$. Finite ladders make the induction terminate;
evaluating at $(S_0,0,\varnothing)$ proves the assertion for $\rho(S_0)$.

(iii) Let every reading be a deterministic function of the own state. Then
$W(s)$ is a number, the reservation equation $(W(s)-g)^+=c(s)$ has the unique
solution $G(s)=W(s)-c(s)$ when a collectable stage remains, $Z(s)=G(s)$, and
$W(s)=\max\{\Yc(s),G(s')\}$ with $s'$ the successor. Consequently an advance
from $s$ to $s'$ either has $\Yc(s)\ge G(s')$, in which case the new incumbent
is at least $\Yc(s)=G(s)+c(s)>G(s)$, or has $\Yc(s)<G(s')$, in which case
$G(s')=G(s)+c(s)>G(s)$; the first case is forced when $s'$ is exhausted.
Suppose $\pi_{\rm IX}$ advances an arm $i\ne j$ at a state where arm $j$, which it has advanced from $S_0$, is not exhausted; then $G_i(s_i)\ge G_j(s_j)$ and $G_i(s_i)>q$.
We claim that at every later state either $q>G_j(s_j)$ or some unexhausted
arm $k$ has $G_k(s_k)>G_j(s_j)$. After the advance of $i$, the first case
above gives $q>G_i(s_i)\ge G_j(s_j)$ and the second gives
$G_i(s_i')>G_i(s_i)\ge G_j(s_j)$. If the claim holds with $q>G_j(s_j)$ it holds
forever, as the incumbent never decreases and $s_j$ does not change while $j$
is paused. If it holds through an arm $k$, the next advance, if any, is of an
arm $m$ with $G_m(s_m)\ge G_k(s_k)>G_j(s_j)$, so $m\ne j$, and the two cases
above restore the claim with $q>G_m(s_m)>G_j(s_j)$ or with $k=m$. In neither
case is $j$ ever index-maximal with index above the incumbent, so it is never
resumed, whatever the tie order.
\end{proof}

\begin{example}[Resumption without learning future rewards]\label{ex:resumption-reward}
Let $q_0=0$, make every stage collectable, and charge $1/10$ per advance.
Arm $A$ first pays zero or one with equal probability, then pays $2/5$
deterministically; independent arm $B$ pays zero or $3/5$ with equal probability
in one stage. The indices are $G_{A,1}=4/5$, $G_B=2/5$, and $G_{A,2}=3/10$.
The index policy starts $A$, switches to $B$ after a zero, and resumes $A$ after
a second zero. Its value is $23/40$, whereas the no-return optimum is $11/20$,
so resumption is worth $1/40$. No reading changes any remaining reward law:
$A$'s index drops at its known stage transition, while the realized collectable
reward determines whether to continue. The calculation below verifies both values.
\end{example}

\paragraph{Value calculations.}
The index policy stops after a first reward of one; after a zero from $A$ it
tries $B$ and completes $A$ only if $B$ also returns zero. Its net utility is
\[
 \tfrac12(1-\tfrac1{10})
 +\tfrac14(\tfrac35-\tfrac2{10})
 +\tfrac14(\tfrac25-\tfrac3{10})=\tfrac{23}{40}.
\]
For a no-return policy starting $A$, a zero leaves the choice between completing
$A$ for additional value $3/10$ and abandoning it for $B$, whose value is $1/5$.
Completing $A$ raises the incumbent to $2/5$, at which inspecting $B$ ties stopping.
Thus starting $A$ has no-return value
$-1/10+(1/2)\cdot1+(1/2)\cdot(3/10)=11/20$.
Starting $B$ and then acting optimally on $A$ gives $21/40$, and stopping gives zero;
these exhaust the first actions and prove the no-return optimum.

\section{Supporting Results for Shared Laws}
\label{app:sharing}

\subsection{Own-arm-law limits and conditional stopping}

\begin{proof}[Proof of Proposition~\ref{prop:no-local}]
Let $q_0=0$. Arm $A$ has one collectable reward $R\sim\operatorname{Bernoulli}(1/2)$ at cost
$0.4$. Arm $B$ has a noncollectable signal $U\sim\operatorname{Bernoulli}(1/2)$ and a
collectable reward identically zero; each of its stages costs $0.01$. In the first instance
$U$ and $R$ are independent; in the second $U=R$. The separate arm laws are identical at the
root, with $G_A=0.2$ and $G_B=-0.02$ in both instances.
In the independent instance, starting $A$ has value $0.5-0.4=0.10$. Starting $B$ costs $0.01$
before the same optimal $A$ decision, giving $0.09$; its zero terminal reward is never worth
buying. In the shared instance, after observing $U=h$ the only potentially profitable action
is to inspect $A$, with net return $h-0.4$. Thus
$Q^\star(B)=-0.01+\E[\max\{0,R-0.4\}]=0.29$ exactly. Starting $A$ still has value $0.10$:
after its reward is known neither stage of $B$ can improve the incumbent or supply useful
information. The value $G_B=-0.02$ also follows from
$\Phi_B(r)=\max\{-0.01,(0-r)^+-0.02\}$ for a run restricted to $B$.
Stopping gives zero. Thus the unique optimal first actions are $A$ and $B$, respectively.
A rule with the stated inputs must choose the same action distribution in both, but optimality
requires probability one on different actions.
\end{proof}

For the general operational planning law, write
$\mathcal R=\{S:\max_{i\in\cI(S)}G_i(S)\le q(S)\}$,
with $\max\varnothing=-\infty$, for the arm-local index stopping region.
\begin{proposition}[Continuation and conditional stopping]\label{prop:stop-any}
Under Assumption~\ref{assm:standing}, at every $S\notin\mathcal R$,
\[
V^\star(S)\ge h(S)+\max_{i\in\cI(S)}\Phi_i^{(\nu_i)}(S;q(S))>h(S).
\]
If $\mathcal R$ is \emph{closed forward}, meaning that $F(S,i,Y)\in\mathcal R$ almost surely
for every $S\in\mathcal R$ and available $i$, then $V^\star(S)=h(S)$ on $\mathcal R$.
Under this additional hypothesis, $\mathcal R$ is exactly the optimal stopping region.
\end{proposition}
\begin{proof}
An attaining single-arm rule exists by finite-horizon backward induction.
Executing it and then stopping is admissible, with value
$h(S)+\Phi_i^{(\nu_i)}(S;q(S))$, proving the first assertion.
For the second, induct on the total stages remaining. At a state in $\mathcal R$, closure and
the induction hypothesis give $V^\star(S')=h(S')$ after every possible first advance.
Therefore
\[
Q^\star(S,i)=h(S)+\E[(\Yc-q(S))^+\mid S,i]-c(S,i)
\le h(S)+\Phi_i^{(\nu_i)}(S;q(S))\le h(S).
\]
The Bellman equation proves stopping optimal; the exhausted-state base case is immediate.
\end{proof}

\begin{example}[A premature index stop]\label{ex:shared-stop}
In the construction above, change $A$'s cost to $0.6$. Its index is now $-0.1$, while $B$'s
remains $-0.02$. The arm-local index stops at value zero, but buying $U$ and inspecting $A$
only when $U=1$ has value $-0.01+\tfrac12(1-0.6)=0.19$. Forward closure fails: after that
signal equals one, $A$'s index is $0.4>q_0$.
\end{example}

Closure quantifies over every available advance in $\mathcal R$, including those
the index never takes. It is sufficient, not necessary. Independent arms have
the exact stopping rule without it: an arm's own observation may raise its index
after an unprofitable initial inspection. The positive-run assertion concerns a
specified single-arm rule, rather than the utility of repeatedly recomputing and
switching under a general shared law. We do not establish closure for general
shared surrogates, including correlated Gaussian-process models.

\subsection{Proof of Proposition~\ref{prop:switching}}
\label{app:switching}

\begin{example}[The switching family]\label{ex:switching}
Let $\Theta\in\{0,1\}$ have prior $p=\Prob(\Theta=1)$ and let arms $A,B$, each with two
stages, be conditionally independent given $\Theta$. Arm $i$ has terminal reward
$R_i\in\{0,1\}$ with $\Prob(R_i=1\mid\Theta=0)=\alpha_i$ and $\Prob(R_i=1\mid\Theta=1)=1/5$,
where $\alpha_A=4/5$ and $\alpha_B=7/10$, and first reading $u_iS_i$, where $S_i\in\{0,1\}$
equals $R_i$ with probability $3/4$ given $R_i$, $u_A=2/5$ and $u_B=3/5$. First stages cost
$1/100$, terminal stages cost $c\in[19/100,1/5]$, and $q_0=0$. Arm $A$'s terminal reward is
more likely to be one when $\Theta=0$, and equally likely when $\Theta=1$;
arm $B$ has the larger intermediate score. Given the terminal rewards,
the signal errors are independent of $\Theta$ and of each other.
The thresholds are
\[
 \tau_1=\frac{15}{19},\qquad
 \tau_2(c)=\frac{132-420c}{50+35c},\qquad
 \tau_3(c)=\frac{288-780c}{250-535c}.
\]
\end{example}

\paragraph{A fixed-cost member.}
At $c=1/5$, the thresholds in Example~\ref{ex:switching} reduce to
$15/19$, $16/19$ and $12/13$. Terminal rewards are always collectable, and the
optimal first advances are
\begin{center}
\renewcommand{\arraystretch}{1.3}
\begin{tabular}{lcccc}
\toprule
Prior $p$ & $(0,\frac{15}{19})$ & $(\frac{15}{19},\frac{16}{19})$
 & $(\frac{16}{19},\frac{12}{13})$ & $(\frac{12}{13},1)$ \\
\midrule
$B$'s signal collectable & $A$ & $B$ & $A$ & $B$ \\
$B$'s signal not collectable & $A$ & $A$ & $A$ & $A$ \\
\bottomrule
\end{tabular}
\end{center}
Either choice of $A$'s intermediate collection rights gives the same rows.

We prove Proposition~\ref{prop:switching} for this family. Write $h=(h_A,h_B)$ for the observed history (each arm has an empty prefix, two
signal prefixes and four completed prefixes, so there are $49$ joint histories
$h$). Let $L_\theta(h)$ be the probability of $h$ given $\Theta=\theta$; every
one of the $16$ joint paths has positive probability in both latent states, so
no history reveals $\Theta$. For a fixed deterministic contingent policy, its
unnormalized payoff from $h$,
\[
 W^\pi_h(p,c)=\sum_\theta \Prob(\Theta=\theta)L_\theta(h)\,
 \E_\theta\Bigl[q(S_\tau)-\textstyle\sum_{t<\tau}c(S_t,i_t)\Bigm| h\Bigr],
\]
is affine in $p$ and affine in $c$, because the incumbent does not depend on
$c$ and each terminal stage contributes $-c$ times a path probability. The
unnormalized optimal value $W_h=\max_\pi W^\pi_h$ is therefore, for fixed $c$,
continuous and piecewise affine in $p$, and satisfies
\[
 W_h=\max\Bigl\{q(h)M_h,\;
 \max_{i}\Bigl[-c_i(h)M_h+\textstyle\sum_{y}W_{(h,i,y)}\Bigr]\Bigr\},
 \qquad M_h=(1-p)L_0(h)+pL_1(h),
\]
with $c_i(h)\in\{1/100,c\}$. Normalizing by $M_h>0$ gives the Bellman equation
for the posterior value, so the optimal first actions at the root are the
maximizers of the bracketed terms at $h=\varnothing$.

\emph{1. Certify all priors at a fixed cost.}
For rational $c$, compute the piecewise-affine Bellman recursion in rational
arithmetic. Partition $[0,1]$ at the union of breakpoints of each state's value,
action values and child values. On each interval, every expression is affine
in $p$: domination of all actions at both endpoints, with one action attaining
the value at both, proves the Bellman identity throughout. Backward induction
therefore certifies the optimum for every prior, rather than only sampled priors.
For example, with both intermediate readings collectable and $c=1/5$, the root
advance values with optimal continuation satisfy
\[
 Q_A(p)-Q_B(p)=\frac{15-19p}{2000},
 \qquad p\in[9/13,63/76].
\]
Their equality gives $\tau_1=15/19$ and the first change from $A$ to $B$;
the certificate also checks the winning advance against stopping.

\emph{2. Certify open intervals of costs.}
Run the same recursion and Bellman certificate over $\mathbb Q(c)$.
Every sort, breakpoint test and certificate comparison depends on the sign of
a rational-function difference $d$. Identically zero differences are exact
identities; for each nonzero $d$, record its sign at a generic rational sample
cost. If none has a zero or pole on an open interval $I$, every comparison
keeps its outcome there and evaluation commutes with the arithmetic.
Thus every $c\in I$ follows the same computational branch and inherits the
all-prior Bellman certificate. Exact Sturm counts on numerators and
denominators establish this sign invariance.

\emph{3. Cover exceptional costs and boundaries.}
On $(19/100,1/5)$, the only interior failure of sign invariance has numerator
proportional to $21000c^2-12035c+1536$, whose root is
$c^\ast=(2407-\sqrt{632689})/8400\approx0.19186$.
Here two partition points used by the computation coincide. This changes the
computation's partition, while the symbolic value and action functions it returns
are identical on the two sides. On $(19/100,c^\ast)$ and $(c^\ast,1/5)$,
all recorded nonzero differences keep their signs and the output gives the
thresholds in Example~\ref{ex:switching}. Fixed-cost certificates cover
$c=19/100$, $c=1/5$ in rational arithmetic and $c=c^\ast$ in
$\mathbb Q(\sqrt{632689})$.
With only $B$'s intermediate reading collectable, the same split and endpoint
checks give the same regions. With terminal-only collection or only $A$'s
intermediate reading collectable, no split is needed: $A$ is uniquely optimal
throughout $(0,1)$ for every cost in the closed interval.

Exact root counts also establish positive threshold denominators and
$0<\tau_1<\tau_2(c)<\tau_3(c)<1$ throughout the cost interval.
For uniqueness, within each open first-action region the winning advance strictly
exceeds stopping and the other advance at every contained breakpoint and midpoint;
nonnegative endpoint differences and positive midpoint differences of affine
functions give strict inequalities in the open piece. The advances tie at
$p=\tau_k(c)$. The verification script and output are included with the code.

\subsection{Signed correction and robustness proof}
\label{app:sharing-robustness}

Throughout this subsection impose the coherent-law and arm-local-cost restriction
of Proposition~\ref{prop:sharing-robustness}. Let
$\mathsf Q_S=\bigotimes_j \mathsf P_{j,S}$ and define
\[
 \mathcal V_\otimes(S)=\E_{\mathsf Q_S}\max\{q(S),Z_1^S,\ldots,Z_N^S\}.
\]
This lies in $[y_{\min},y_{\max}]$ and is the independent-surrogate optimum,
rather than the true shared optimum. Use the continuation utility
$U_\pi=V_\pi+K$ defined in Section~\ref{sec:setting}.

After forcing arm $i$ and seeing $Y=y$, write $S'=F(S,i,y)$. Define
$\mathcal V_i^{\rm fr}(S,y)$ using the new incumbent and the advanced arm's
conditional remaining law, but retain $\mathsf P_{j,S}$ for $j\ne i$. Coherence ensures
the advanced law equals its actual marginal at $S'$; only untouched laws differ.
Put
\begin{align}
 d_0(S)&=\mathcal V_\otimes(S)-q(S),\nonumber\\
 d_i(S)&=\mathcal V_\otimes(S)+c(S,i)-\E[\mathcal V_i^{\rm fr}(S,Y)\mid S,i],\nonumber\\
 \Gamma_i(S)&=\E[\mathcal V_\otimes(S')-\mathcal V_i^{\rm fr}(S,Y)\mid S,i].
 \label{eq:sharing-gamma}
\end{align}
Here $d_0$ is the independent surrogate's stop loss, $d_i$ its loss from
advancing $i$ under the frozen update, and $\Gamma_i$ the signed change in
the surrogate potential from updating untouched arms. Both signs of $\Gamma_i$
are possible; the proof controls its magnitude.
The losses $d_0,d_i$ are nonnegative by Theorem~\ref{thm:loss} applied to
$\mathsf Q_S$. In particular, with
$O_i=\max\{q,\max_{j\ne i}Z_j^S\}$ independent of arm $i$ under $\mathsf Q_S$,
\[
 d_i=c(S,i)-\E_{\mathsf Q_S}[(W_i^S-\max\{O_i,G_i(S)\})^+].
\]
For finite $G_i$ this is the capped-overshoot expression of Theorem~\ref{thm:loss};
for $G_i=-\infty$ it equals $c(S,i)$.

\begin{proof}[Proof of Proposition~\ref{prop:sharing-robustness}]
The true one-step residual is the signed update correction less the
nonnegative surrogate action loss:
\begin{equation}
 -c(S,i)+\E_{\mathsf P}[\mathcal V_\otimes(S')\mid S,i]-\mathcal V_\otimes(S)
       =\Gamma_i(S)-d_i(S).
 \label{eq:sharing-residual}
\end{equation}
Summing conditional expectations with indicators before the stopping time over
the deterministic range $0,\ldots,H-1$ gives, for every policy,
\begin{equation}
 U_\pi(S_0)=\mathcal V_\otimes(S_0)
 +\E_\pi\sum_{b<\tau_\pi}(\Gamma_{i_b}-d_{i_b})(S_b)
 -\E_\pi d_0(S_{\tau_\pi}).
 \label{eq:sharing-identity}
\end{equation}
All sums are finite and bounded. At each chosen index advance the surrogate
gap $d_i$ is zero; at its stop $d_0=0$. Consequently
\begin{equation}
 U_{\pi_{\rm IX}}(S_0)=\mathcal V_\otimes(S_0)
       +\E_{\pi_{\rm IX}}\sum_{b<\tau_{\rm IX}}\Gamma_{i_b}(S_b).
 \label{eq:sharing-index-identity}
\end{equation}
Subtracting from~\eqref{eq:sharing-identity} for a true optimal policy
$\pi^\star$, dropping its nonnegative stop gap, and taking positive parts yields
\begin{equation}
 U^\star-U_{\pi_{\rm IX}}
 \le\E_{\pi^\star}\sum_{b<\tau_\star}(\Gamma_{i_b}-d_{i_b})^+(S_b)
   +\E_{\pi_{\rm IX}}\sum_{b<\tau_{\rm IX}}\Gamma_{i_b}^-(S_b),
 \label{eq:sharing-refined}
\end{equation}
where $x^-=(-x)^+$. The first sum allows positive corrections along the true
optimal trajectory; the second allows negative corrections along the index
trajectory. This is why both trajectories enter the comparison.

Fix $S,i,y$. Replace the frozen untouched-arm capped distributions by their
updated distributions one arm at a time. For each replacement, condition on
all other independent capped variables. Their maximum with the new incumbent
is an outside option in $[q(S'),y_{\max}]$, so the absolute value change is at
most the corresponding supremum defining $\varepsilon_i(S)$. The advanced
arm's law is common to both products. Triangle inequality and expectation give
$|\Gamma_i(S)|\le\varepsilon_i(S)$. Each summand
in~\eqref{eq:sharing-refined} is at most $\bar\varepsilon$ and each trajectory
has at most $H$ advances. The two contributions are therefore at most
$H\bar\varepsilon$ each, giving $2H\bar\varepsilon$; removing the same sunk
cost from both values does not change regret.
\end{proof}

This is the finite-horizon approximate-planning argument. In the formulation of
\citet[Theorem~9 and Remark~11]{subramanian2022ais}, use the current marginal
laws and own histories as the abstract state, frozen updates as the approximate
transition, and incremental reward $q(S')-q(S)-c(S,i)$. The approximate value
is $\mathcal V_\otimes-q$ and its expected next-value discrepancy is exactly
$\Gamma_i$. The envelope above specializes that discrepancy using local
stop-loss values. The local values are $1$-Lipschitz in the outside option,
so their supremum is measurable by using a countable dense set and the
endpoints. Replacing $Z_j$ by $\max\{y_{\min},Z_j\}$ handles exhausted arms
without infinite arithmetic.

\paragraph{A path-total-variation envelope.}
Let $D=y_{\max}-y_{\min}$ and put $B_j(S)=D+\nu_j(S)c_{\max}$ if a
collectable stage remains, and zero otherwise. With
$\operatorname{TV}(\mu,\nu)=\sup_A|\mu(A)-\nu(A)|$,
\begin{equation}
 \varepsilon_i(S)\le\varepsilon_i^{\rm TV}(S)
 =\E\!\left[\sum_{j\ne i} B_j(S)
   \operatorname{TV}(\mathsf P_{j,S'},\mathsf P_{j,S})\,\middle|\,S,i\right].
 \label{eq:sharing-tv}
\end{equation}
With densities, our TV is half the $L^1$ distance; the convention in
\citet[Section~3.1]{subramanian2022ais} is the full $L^1$ distance.
The constants here use our stated convention throughout.
Indeed, each fixed local stopping rule, including immediate stopping, has path
payoff $\max\{r,M_\sigma\}-\kappa_\sigma$ in
$[y_{\min}-\nu_jc_{\max},y_{\max}]$. The same rule class is feasible under
both marginal path laws because costs are arm-local. TV bounds each expectation
difference by its payoff oscillation times TV, and taking the supremum over
rules preserves the bound. Without a collectable stage, immediate stopping is
optimal under both laws. This proves~\eqref{eq:sharing-tv}.

\begin{proof}[Proof of Corollary~\ref{cor:component}]
Conditional independence given $\Theta$ makes arm $j$'s remaining-path law at
$S$ the mixture $\int\mathsf P_j^\vartheta(\cdot\mid s_j)\,\mu_S(d\vartheta)$; a reading on
arm $i$ changes the mixing measure to $\mu_{S'}$ and leaves the components
unchanged. Write $\mu_{S'}-\mu_S=\xi^+-\xi^-$ by the Jordan decomposition, with
common mass $t=\operatorname{TV}(\mu_{S'},\mu_S)$. If $t=0$ the untouched
marginal laws agree; otherwise
$\mathsf P_{j,S'}-\mathsf P_{j,S}=t\bigl(\int\mathsf P_j^\vartheta\,d\xi^+/t-\int\mathsf P_j^\vartheta\,d\xi^-/t\bigr)$,
and the total variation between two mixtures of the components is at most the
largest total variation between components, so
$\operatorname{TV}(\mathsf P_{j,S'},\mathsf P_{j,S})\le\Lambda_j(S)\,t$.
Next, let $J$ be the joint law of $(\Theta,Y)$ given $S,i$, let $\beta$ be
its $Y$ marginal, and put $M=\mu_S\otimes\beta$. The finite measure $J+M$
dominates both $J$ and $M$. Integrating the absolute difference of their
densities with respect to $J+M$, first conditional on $Y$ and then conditional
on $\Theta$, gives the total-variation disintegration identity
\[
 \E_Y\operatorname{TV}(\mu_S(\cdot\mid Y),\mu_S)
 =\operatorname{TV}(J,M)
 =\E_{\vartheta\sim\mu_S}
   \operatorname{TV}(\mathsf P_i^\vartheta(Y\in\cdot\mid s_i),\beta)
 =\eta_i(S).
\]
Regular conditional laws exist on the stated standard Borel spaces.
This argument applies also to singular joint laws, without assuming $J\ll M$.
Substituting into~\eqref{eq:sharing-tv} gives the envelope, and
Proposition~\ref{prop:sharing-robustness} the regret bound. The two-draw bound
is the convexity inequality
$\operatorname{TV}(\mathsf P^\vartheta,\int\mathsf P^{\vartheta'}\mu(d\vartheta'))\le\int\operatorname{TV}(\mathsf P^\vartheta,\mathsf P^{\vartheta'})\mu(d\vartheta')$.
The binary formulas are the case of two components, computed next, in which $\varepsilon_i^{\rm TV}$ equals the displayed product.
\end{proof}

If arm paths are independent conditional on binary $\Theta$, let
$p=\mathsf P(\Theta=1\mid S)$, and let $D_i^{\rm next}$ and $D_j^{\rm path}$ be the
TV separations of the next-reading and remaining-path laws, respectively,
between the two latent components. Then, for $0<p<1$,
\begin{equation}
 \varepsilon_i^{\rm TV}(S)
 =2p(1-p)D_i^{\rm next}(S)\sum_{j\ne i} B_j(S)D_j^{\rm path}(S).
 \label{eq:sharing-binary}
\end{equation}
Here $\Theta$ is an uncertain latent state under the fixed planning law,
distinct from its fixed surrogate parameters $\theta$.
To see this, set $p_y=\mathsf P(\Theta=1\mid S,Y_i=y)$. An untouched arm's component
laws are unchanged, so its mixture-TV shift is
$|p_y-p|D_j^{\rm path}$. Bayes' rule gives
$\E|p_Y-p|=2p(1-p)D_i^{\rm next}$ by taking absolute variation of
$(p_y-p)\mu(dy)=p(1-p)(\mu_1-\mu_0)(dy)$ for
$\mu=(1-p)\mu_0+p\mu_1$. At $p\in\{0,1\}$ the envelope is zero, without
defining null-component conditionals.

\paragraph{Concentration and predictive dispersion.}
Weak posterior concentration alone does not force $\eta_i$ to vanish.
For example, consider beliefs $\mu_n=\operatorname{Uniform}[0,1/n]$ on a common
component $\Theta$, with every arm's next reading equal to $\Theta$.
These degenerate paths are conditionally independent given $\Theta$ and have
bounded readings. Although $\mu_n$ converges weakly to a point mass at zero,
$\operatorname{TV}(\delta_\vartheta,\mu_n)=1$ for every $\vartheta$, so
$\eta_i=1$ for every $n$. Predictive total-variation regularity is needed to
deduce local dispersion decay from concentration in a continuous component
space. To deduce a vanishing regret bound, the envelope must additionally
vanish uniformly over feasible future states, as required in
Corollary~\ref{cor:component}.

\paragraph{Why future-state control is required.}
Let $(U,A,B)$ be uniform over $000,011,101,110$. Each pair is independent.
Inspecting noncollectable $U$ costs $1/100$; inspecting collectable $A$ or $B$
costs $3/5$ each, with $q_0=0$. Every root envelope is zero because one root
observation leaves each other marginal unchanged, and the index stops.
After $U=1$, $A$ and $B$ are opposite: inspect $A$, then $B$ only if $A=0$,
for value $1/10$. After $U=0$, stopping is optimal. Buying $U$ therefore has
value $1/25$, whereas starting $A$ or $B$ has value $-1/200$. Later states have
nonzero envelopes. Thus even pairwise independence of whole arm paths at the
root cannot replace the all-history condition.

\clearpage
\section{Numerical Verification}
\label{app:experiments}

The checks use fully specified finite laws, exact rational arithmetic, and a
joint Bellman solver independent of the arm-wise covered-call recursion.
Within each instance, policies use the same law, costs, collection rights and
observation protocol. Table~\ref{tab:verification-summary} summarizes the
verification; each study's model, checks and results are given together below.

\begin{table}[htb]
\centering\small
\begin{tabular}{p{0.25\linewidth}rp{0.53\linewidth}}
\toprule
Instance family & Cases & Exact checks \\
\midrule
Independent trees & 192 & Value representation, maximal-index actions, stop and advance losses \\
Staging grid & 660 & Index optimality and accumulated policy losses \\
Comparative pairs & $1763+60$ & Coupled covered-call overshoot identity on grid and random pairs \\
Shared-law audit & 453 & Signed residual, policy telescoping, and local-value/path-TV envelopes \\
Concentration sweep & 240 & Policy accounting and sharing envelopes at five binary-component priors \\
\bottomrule
\end{tabular}
\caption{All checks pass in exact arithmetic against direct Bellman calculations.
Cases count starting instances, grid cells or problem pairs as indicated;
they are verification coverage, not independent statistical observations.}
\label{tab:verification-summary}
\end{table}

\subsection{Independent-arm verification}
\label{app:finite-results}

Two independent binary-tree arms are generated in four collection regimes:
every stage on depth three, terminal only on depth four, $\{1,3\}$ on depth
three, and $\{2,4\}$ on depth four. At each own history, independently choose
a cost from $\{1/100,3/100,3/50,1/10\}$, two distinct readings from
$\{0,1/4,1/2,3/4,1\}$, and a branch probability from $\{1/4,1/2,3/4\}$.
Each regime uses seeds $2026102400$ through $2026102423$ and is evaluated from
both the empty state and a prefix after three random feasible advances,
giving 192 starting instances. Prefixes are chosen without an index or
stopping criterion and retain their observed incumbent.

The covered-call value, maximal-index actions above the incumbent, and all
stop and advance losses agree with direct Bellman optimization over
$23{,}839$ evaluated states and $37{,}467$ available advance actions.

\subsection{Staging and policy-loss verification}
\label{app:staging-results}

Each independent arm reveals two noisy signals followed by a fair binary
terminal reward $H_i$. Conditional on $H_i$, signals $S_{i1},S_{i2}$ are
independent and equal $H_i$ with probability $\alpha$. The readings are
$(2S_{i1}/5,3S_{i2}/5,H_i)$, and predictions condition on both signals.
The grid crosses
\[
\begin{split}
 \alpha&\in\{1/2,5/8,3/4,7/8,1\},\qquad
 k\in\{1/200,1/50,1/20,1/10\},\\
 c&\in\{1/20,1/5,2/5,3/5,4/5\},\qquad
 \cK_i\in\{\{3\},\{1,3\},\{1,2,3\}\}.
\end{split}
\]
With arm multipliers $m_i\in\{1,5/4\}$, the constant-cost family uses
$m_i(k,k,c)$ and the own-history-cost family uses
\[
 m_i\left(k,\;k(1+S_{i1}/2),\;
 c\left(1+\tfrac12\mathbf 1\{S_{i1}\ne S_{i2}\}\right)\right).
\]
Each next cost is known before the advance. Deterministic controls have readings
$(0,1/2,1)$ and $(0,3/8,3/4)$, costs $m_i(k,k,c)$, and the same cost and
collection grids. There are 100 constant-cost cells, 100 own-history-cost cells
and 20 deterministic controls per collection regime, giving 660 cells.
All runs start at $q_0=0$; zero-probability prefixes at $\alpha=1$ are omitted.
The index stops on an index--incumbent tie and breaks advance ties by the smaller label.

Three specified deviations check Theorem~\ref{thm:loss}.
\emph{Stop after one} takes the initial index advance and stops, or stops immediately
if the index does. \emph{Lowest index first} forces a smallest-index available arm
once, even if stopping was optimal, then follows the index.
\emph{Extra at stop} follows the index to its first stop, takes one cheapest
available advance if any, and then follows the index. All action ties favor
the smaller label. Each deviation's utility loss equals its accumulated stop
and advance losses in every cell. The index agrees with the Bellman optimum
over $48{,}840$ state evaluations and $74{,}160$ advance-action checks,
including all 300 own-history-cost cells.

\subsection{Comparative statics and resumption}
\label{app:comparative-results}

A dominated transition pairs staging cells with the same path laws and arms,
but the next larger collectable set, next lower signal cost, next lower terminal
cost, or constant in place of own-history costs. Coupling each arm's capped
values verifies Proposition~\ref{prop:comparative} on all $1{,}763$ transitions.
The identity also holds on 60 random binary-tree pairs of depths two to four
(seed $20261030$), with enlarged collection sets and costs multiplied by
$1$, $1/2$ or $3/4$, including unchanged cases.

Table~\ref{tab:comparative} summarizes the three checks by transition type,
collectable set and cost law.
For Proposition~\ref{prop:fewer-rights}, terminal-only indices are used under
intermediate collection in every cell with intermediate rights. The policy
attains the expanded-rights optimum in nearly half of these cells and, in about
a quarter, loses the full value of the additional rights, so both ends of the
proposition's bound are attained. At $\alpha=1/2$, $k=1/200$ and $c=3/5$ with
$\cK_i=\{1,3\}$, for instance, the terminal-only indices are negative, the
policy stops at once, and its loss is the full rights value.
For Proposition~\ref{prop:resumption}, a separate Bellman recursion forbids
further advances of any arm once another arm is advanced after it, while
retaining all collectable readings. This gives $V_{\rm NR}$.
Resumption has positive value exactly in the stochastic cells where the index
policy resumes, and in no deterministic control; the largest gap is under one
twentieth of its cell's optimal value. The bound $\rho$ never exceeds the gap,
is positive in the same cells, and attains the gap in about a fifth of them.

\begin{table}[htb]
\centering\small
\begin{tabular}{lrr}
\toprule
Dominated transition & Pairs & Identity holds \\
\midrule
Larger collectable set & 440 & 440 \\
Lower signal cost & 495 & 495 \\
Lower terminal cost & 528 & 528 \\
Constant for own-history costs & 300 & 300 \\
\midrule
Collectable set & Cells & Zero loss / full rights value lost \\
\midrule
$\{1,3\}$ & 220 & 133 / 50 \\
$\{1,2,3\}$ & 220 & 70 / 51 \\
\midrule
Cost law & Cells & Positive gap / $\rho$ equal to gap / largest gap \\
\midrule
Constant & 300 & 76 / 17 / 0.0108 \\
Own-history & 300 & 59 / 13 / 0.0156 \\
Deterministic controls & 60 & 0 / 0 / 0.0000 \\
\bottomrule
\end{tabular}
\caption{Comparative statics and resumption on the 660-cell staging grid, in
exact arithmetic. Top: dominated pairs of cells; the value difference equals
the covered-call overshoot of Proposition~\ref{prop:comparative} in every pair.
Middle: terminal-only indices under intermediate collection; the loss lies in
$[0,V^{\star\prime}-V^\star]$, and the last entry counts cells with positive
rights value in which all of it is lost. Bottom: the no-return optimum; the gap
$V^\star-V_{\rm NR}$ is positive exactly where the index policy resumes, and
$\rho$ never exceeds it.}
\label{tab:comparative}
\end{table}

\subsection{Shared-law verification}
\label{app:sharing-results}

The audit has 432 two-stage binary-latent cells, 12 arbitrary three-stage joint
laws with own-history costs, and nine settings of the constructed example in
Appendix~\ref{app:sharing-controls}. All start with unobserved arms and $q_0=0$.
The joint Bellman solver retains dependence; the index recomputes from the true
posterior arm marginals after every observation.

\paragraph{Binary-latent grid.}
Conditional on $\Theta\in\{0,1\}$, complete arm paths are independent.
Arm $i$ has $R_i\sim\operatorname{Bernoulli}(a_{i\Theta})$ and first signal
$S_i$ satisfying $\Prob(S_i=R_i\mid R_i,\Theta)=r_i$. Its readings are
$(s_iS_i,R_i)$, with $(s_A,s_B)=(2/5,3/5)$. The terminal is collectable;
each first stage can independently be designated collectable, giving four choices.
Seed $20261026$ generates 12 base models with prior $p_0\in\{1/5,1/2,4/5\}$.
For each arm choose two distinct component probabilities from
$\{1/10,1/5,3/10,1/2,7/10,4/5,9/10\}$, $r_i\in\{3/5,3/4,9/10\}$,
first cost in $\{1/100,1/25,2/25\}$ and terminal cost in $\{1/20,1/5,2/5\}$.
For each model set
\[
\begin{gathered}
 m_i=(1-p_0)a_{i0}+p_0a_{i1},\qquad
 a_{i\theta}(\lambda)=m_i+\lambda(a_{i\theta}-m_i),\\
 \lambda\in\{0,1/100,1/50,1/20,1/10,1/5,2/5,3/5,1\}.
\end{gathered}
\]
This holds the full initial arm marginal laws, costs and collection choices
fixed within each coupling sweep. At $\lambda=0$ the arms are independent,
giving 48 independent controls among the $12\times9\times4=432$ cells.

\paragraph{Arbitrary three-stage laws.}
Continuing the same seeded generator, twelve models assign normalized positive
integer weights from $\{1,\ldots,9\}$ to all 64 binary joint paths of two
three-stage arms. Base costs are $(1/20,1/10,1/5)$; the second cost is $1/4$
after first reading one, and the terminal cost is $1/25$ after prefix $(0,1)$.
For model $b\in\{0,\ldots,11\}$ and arm $i\in\{0,1\}$, stage one is
collectable when $(b+i)\bmod2\ne0$, stage two when $(b+i)\bmod3\ne0$, and
stage three always. These laws test sharing without assuming the binary-latent factorization.

\paragraph{Checks and results.}
Every feasible prefix and available advance is checked, including histories
the index never visits. The index stops on an index--incumbent tie and breaks
advance ties by smaller label. The Bellman policy used for the refined bound
favors advances over stopping, then smaller label; this fixes its occupancy at
a value tie. Policy telescoping is checked for the index, the true optimum,
immediate stopping, and a smallest-label always-advance policy.
Uniform envelopes maximize over all feasible histories; the refined bound
uses optimal-policy occupancy and is an oracle diagnostic.
Across the 453 cells, the signed residual, telescoping and call/TV inequalities
hold over $24{,}309$ feasible prefixes and $21{,}042$ available advances.
The independent controls have zero sharing envelopes and zero regret.
Negative signed corrections occur, supporting the two-sided accounting
in~\eqref{eq:sharing-refined}. Counts measure verification coverage;
these known-law checks do not assess learned-surrogate accuracy or practical
optimization performance.

\subsection{Positive regret, shared nuisance and concentration}
\label{app:sharing-controls}

\paragraph{Positive-regret example.}
Use the binary-latent construction above with $p_0=4/5$,
$(a_{A0},a_{A1})=(4/5,1/5)$, $(a_{B0},a_{B1})=(7/10,1/5)$,
$r_A=r_B=3/4$, first scores $(2/5,3/5)$, first costs $(19/500,1/20)$,
and terminal costs $(1/5,1/5)$. Both stages are collectable, $q_0=0$,
and the same nine coupling settings are included in the audit.
Both initial indices are $1/2$; ties favor $A$.
The positive regret in Table~\ref{tab:robustness-bounds} demonstrates the loss
of exact optimality under sharing, while each displayed bound covers that regret.

\paragraph{Shared nuisance.}
Outside the 453-cell audit, let the paths be $(U,R_A)$ and $(U,R_B)$, where
$U$ is a shared fair bit and $R_A,R_B$ are independent
$\operatorname{Bernoulli}(4/5)$ rewards, also independent of $U$.
First readings are noncollectable at cost $1/100$; terminals are collectable
at cost $1/5$. The shared bit changes predictions of another noncollectable
reading but leaves local continuation values unchanged.
The call envelope is zero, the index is optimal with utility $177/250$, and
the uniform path-TV bound is $28/5$. Root joint/product TV is $1/2$, a different
quantity from the posterior marginal shifts in the path-TV envelope.
The control isolates why local-value changes, rather than all remaining-path
changes, are the discrepancy relevant to the bound.

\paragraph{Concentration sweep.}
Also outside the audit, the 12 base models are reused at full component
contrast ($\lambda=1$) with all four collection choices, and only the prior
$p=\Prob(\Theta=1)$ is varied over $\{1/2,1/4,1/10,1/40,1/160\}$: 240 cells
in which the belief about the binary component concentrates while the
component laws, reliabilities, costs and collection stay fixed, each running
the same exact audit as the grid. For this binary component, $\eta_i$ at the
root is $2p(1-p)$ times the component separation of the next reading
(Corollary~\ref{cor:component}). Table~\ref{tab:concentration} reports, for
each prior, the largest regret and the largest uniform call and path-TV bounds
over its 48 cells. Among the tested priors, regret is positive only at $p=1/2$,
and the largest uniform bounds fall by more than an order of magnitude from
$p=1/2$ to $p=1/160$. The bounds remain conservative. Their decrease is an
observed property of this prior-sensitivity calculation on fixed finite laws;
the corollary requires control of dispersion over all feasible future histories,
in addition to concentration at the root.

\begin{table}[htb]
\centering\small
\begin{tabular}{lrrrrr}
\toprule
Prior $p$ & $2p(1-p)$ & Positive-regret cells & Max regret & Max call bound & Max path-TV bound \\
\midrule
$1/2$ & 0.5000 & 4 & 0.00698 & 0.896 & 2.015 \\
$1/4$ & 0.3750 & 0 & 0.00000 & 0.715 & 1.680 \\
$1/10$ & 0.1800 & 0 & 0.00000 & 0.446 & 1.070 \\
$1/40$ & 0.0488 & 0 & 0.00000 & 0.166 & 0.398 \\
$1/160$ & 0.0124 & 0 & 0.00000 & 0.046 & 0.111 \\
\bottomrule
\end{tabular}
\caption{Concentration sweep over the prior of the binary component: the 12
base models at full contrast and all four collection choices, 48 cells per row.
Regret and the uniform bounds are row maxima; each cell's bounds maximize over
all feasible histories. The factor $2p(1-p)$ describes the root belief in
Corollary~\ref{cor:component}. Exact values are rounded.}
\label{tab:concentration}
\end{table}

\begin{table}[htb]
\centering\small
\begin{tabular}{lrrrr}
\toprule
Instance & Regret & Refined (oracle) & Call & Path-TV \\
\midrule
Witness, $\lambda=1/10$ & 0.000024 & 0.000156 & 0.002142 & 0.012600 \\
Witness, $\lambda=1$ & 0.001700 & 0.014900 & 0.214158 & 1.260000 \\
Shared nuisance & 0.000000 & 0.000000 & 0.000000 & 5.600000 \\
\bottomrule
\end{tabular}
\caption{Checks of the sharing bound under specified joint laws. Refined uses
the true optimal policy occupancy in~\eqref{eq:sharing-refined}; Call and Path-TV
are uniform $2H\bar\varepsilon$ bounds. Values are rounded to six decimals.
The nuisance control is outside the 453-cell audit.}
\label{tab:robustness-bounds}
\end{table}

\section{Index Computation}
\label{app:computation}

\subsection{Conditional-law calculations}
Algorithm~\ref{alg:staged-index} constructs each covered-call law by backward
recursion. For discrete $W$, sorting its finite atoms and sweeping their
cumulative mass and first moment solves the piecewise-linear reservation
equation exactly. These calculations integrate over hypothetical futures;
an actual advance does not disclose $W$ or $Z$.

Continuous reading laws require integration or approximation; no
numerical-integration error guarantee is added here.

\subsection{Finite-tree complexity}
An arm with $T$ remaining stages and $J$ possible next observations is a depth-$T$ observation
tree. At a node with $t$ observations taken, $W$ has at most $J^{T-t}$ atoms. Sorting them and
sweeping cumulative probability and first moment solves the piecewise-linear reservation
equation exactly for a discrete law. A direct implementation sorting at each node has the
conservative bound $O(T^2 J^T\log(1+J))$ for these atom operations, excluding evaluation of the
predictive kernels. Sorted-list merging can avoid repeated sorting, but no such improvement is
needed for the present guarantees. Here $J\ge1$ is an upper bound on the number of
next observations; using $\log(1+J)$ includes deterministic branches. For $J\ge2$,
the bound is exponential in $T$. The index separates
the arms rather than solving their joint Bellman problem; it does not remove the need to
integrate each arm's possible future histories. The cost of evaluating predictive kernels must also be included when measuring decision time.

\paragraph{Code and archived checks.}
The code, records and verification scripts are at
\url{https://github.com/longvus/staged-pandora-box}; its
\path{next_paper/P2_REPRODUCIBILITY.md} identifies the reference solvers,
protocols, archived records and validation commands.

\end{document}